\documentclass[letterpaper, 10 pt, conference]{ieeeconf}  
\IEEEoverridecommandlockouts
\usepackage{amsmath,amsfonts}
\usepackage{algorithmic}
\usepackage{array}
\usepackage{textcomp}
\usepackage{stfloats}
\usepackage{url}
\usepackage{verbatim}
\usepackage{graphicx}
\def\BibTeX{{\rm B\kern-.05em{\sc i\kern-.025em b}\kern-.08em
    T\kern-.1667em\lower.7ex\hbox{E}\kern-.125emX}}
\usepackage{balance}

\usepackage{amsfonts}
\usepackage{amssymb}
\usepackage{mathtools}
\usepackage{bm}  
\usepackage{graphics} 
\usepackage{epsfig} 
\usepackage{wrapfig}
\usepackage{subcaption}
\usepackage[font=small,labelfont=small]{caption}
\usepackage{color}
\usepackage{algorithm}
\usepackage{algorithmic}
\usepackage{sidecap}
\sidecaptionvpos{figure}{c}
\usepackage{tikz}
\usetikzlibrary{arrows.meta, positioning, backgrounds}
\usepackage{tabularx}
\usepackage{colortbl}
\usepackage{booktabs}
\usepackage{multirow}
\usepackage{siunitx}
\usepackage{nicefrac}
\usepackage[nolist]{acronym} 
\usepackage{cuted}
\usepackage{rotating}
\usepackage{leftindex}
\usepackage[noadjust]{cite}  
\usepackage{svg}
\usepackage{xfrac}
\usepackage{adjustbox}

\usepackage{amsthm} 
\usepackage[framemethod=tikz]{mdframed} 
\usepackage{lipsum}

\usepackage{soul} 
\usepackage[dvipsnames]{xcolor}
\usepackage[normalem]{ulem}

\usepackage{amsmath}

\newcommand{\trsp}{\mathsf{T}}  
\newcommand{\ty}[1]{{\scriptscriptstyle{\mathcal{#1}}}}

\newcommand{\euclideanspace}{\mathbb{R}}

\newcommand{\spd}[1]{\mathcal{S}_{\ty{++}}^{#1}}

\newcommand{\specialeucl}[1]{\operatorname{SE}(#1)}

\newcommand{\configmanifold}{\mathcal{Q}}

\newcommand{\configtangentspace}[1]{\mathcal{T}_{#1}\configmanifold}
\newcommand{\configcotangentspace}[1]{\mathcal{T}^*_{#1}\configmanifold}

\newcommand{\redmanifold}{\check{\configmanifold}}

\newcommand{\redtangentspace}[1]{\mathcal{T}_{#1}\redmanifold}
\newcommand{\redcotangentspace}[1]{\mathcal{T}^*_{#1}\redmanifold}

\newcommand{\group}{\mathbb{G}}

\newcommand{\generallineargroup}[1]{\text{GL}(#1)}

\newcommand{\euclideangroup}[1]{\mathrm{E}(#1)}
\newcommand{\speceuclideangroup}[1]{\mathrm
{SE}(#1)}
\newcommand{\speceuclideanliealgebra}[1]{\mathfrak{se}(#1)}
\newcommand{\orthogonalgroup}[1]{\mathrm{O}(#1)}
\newcommand{\specorthogonalgroup}[1]{\mathrm{SO}(#1)}

\newcommand{\cyclicgroup}[1]{\mathbb{C}_{#1}}

\newcommand{\set}{\mathbb{S}}
\newcommand{\lgroupactionblank}{\triangleright}
\newcommand{\lgroupactiononblank}[1]{\triangleright_{\scriptscriptstyle #1}}
\newcommand{\lgroupactionon}[2]{\triangleright_{\scriptscriptstyle #1}{#2}}

\newcommand{\rgroupactiononblank}[1]{\leftindex_{\scriptscriptstyle #1}\triangleleft}
\newcommand{\rgroupactionon}[2]{{#2}\leftindex_{{\scriptscriptstyle #1}}\triangleleft}

\newcommand{\rep}[1]{\psi_{#1}}

\newcommand{\Gshape}{\group_{\body}}
\newcommand{\Gmass}{\group_{\rho}}
\newcommand{\Gmat}{\group_{W}}
\newcommand{\Gmorph}{\group_{\mathrm{morph}}}
\newcommand{\Genv}{\group_{\mathrm{env}}}
\newcommand{\Gtask}{\group_{\mathrm{task}}}
\newcommand{\Gdisc}{\group_{\mathrm{disc}}}
\newcommand{\Gstar}{\group^{\star}}
\newcommand{\Ggrav}{\group_{\gvec}}

\newcommand{\Dinfh}{D_{\infty h}}

\newcommand{\Fl}{M_z}

\newcommand{\Mx}{M_x}

\newcommand{\q}{\bm q}
\newcommand{\gvec}{\bar{\bm g}}

\newcommand{\body}{\mathcal{B}}

\tikzset{
  sgrp/.style={draw, rounded corners=2pt, inner sep=1.5pt, font=\footnotesize, fill=white, align=center},
  only3d/.style={fill=gray!14},
  chiral/.style={fill=orange!16},
  rare/.style={dotted, thick},
  unused/.style={draw=gray!60, text=gray!70},
  hl/.style={line width=1.1pt},
  cov/.style={-, semithick, black!75},
  covu/.style={-, semithick, gray!45},
  noncov/.style={-, semithick, dashed, gray},
  cause/.style={font=\tiny, text=black!60, align=center},
}
\pgfdeclarelayer{bg}
\pgfsetlayers{bg,main}

\theoremstyle{plain}
\newtheorem{definition}{Definition}

\theoremstyle{plain}
\newtheorem{proposition}{Proposition}

\newtheorem{principle}{Principle}

\usepackage[colorlinks=True,
            linkcolor=blue,
            citecolor=blue,
            urlcolor=blue]{hyperref}

\begin{document}

\title{\LARGE \bf Morphological Symmetries of Deformable Objects: \\ Theory, Taxonomy, and Applications to Robot Manipulation \vspace{-1.5ex}}

\author{
Riccardo Morandi\authorrefmark{1},
Katharina Friedl\authorrefmark{1},
Loizos Hadjiloizou\authorrefmark{2}, 
Francesco Buso\authorrefmark{2}, 
Seungyeon Kim\authorrefmark{2}, \\
Ulises Campod\'{o}nico, 
Danica Kragic, and
No\'{e}mie Jaquier
    \thanks{
    \authorrefmark{1}, \authorrefmark{2}: equal contribution.
    All authors are with the Department of Robotics, Perception, and Learning, KTH Royal Institute of Technology.  Correspondence to: \tt{\footnotesize \{\href{mailto:rmorandi@kth.se}{rmorandi},\href{mailto:kfriedl@kth.se}{kfriedl},\href{mailto:jaquier@kth.se}{jaquier}\}@kth.se}.}
}



\maketitle
\newacro{ae}[AE]{Autoencoder}
\newacro{rom}[ROM]{reduced-order model}
\newacro{fom}[FOM]{full-order model}
\newacro{mor}[MOR]{model order reduction}
\newacro{lnn}[LNN]{Lagrangian neural network}
\newacro{hnn}[HNN]{Hamiltonian neural network}
\newacro{dof}[DoF]{degrees-of-freedom}
\newacro{ivp}[IVP]{initial value problem}
\newacro{dnn}[DNN]{deep neural network}
\newacro{cnn}[CNN]{convolutional neural network}
\newacro{gnn}[GNN]{graph neural network}
\newacro{spd}[SPD]{symmetric positive-definite}
\newacro{mlp}[MLP]{multilayer perceptron}
\newacro{emlp}[EMLP]{equivariant multilayer perceptron}
\newacro{fc}[FC]{fully-connected}
\newacro{gyrospd}[GyroSpd$_{\ty{++}}$]{gyrospace hyperplane-based}
\newacro{gyroai}[GyroAI]{gyrocalculus-based}
\newacro{rolnn}[RO-LNN]{reduced-order Lagrangian neural network}
\newacro{rohnn}[RO-HNN]{reduced-order Hamiltonian neural network}
\newacro{pde}[PDE]{partial differential equation}
\newacro{hnko}[HNKO]{Hamiltonian neural Koopman operator}
\newacro{mpc}[MPC]{model-predictive control}
\newacro{rl}[RL]{reinforcement learning}
\newacro{pd}[PD]{proportional derivative}
\newacro{iss}[ISS]{input-to-state stable}
\newacro{mshgnn}[MS-HGNN]{Morphological-Symmetry-Equivariant Heterogeneous Graph Neural
Network}
\newacro{mslnn}[MS-LNN]{morphologically-symmetric geometric Lagrangian neural network}
\newacro{msrolnn}[MS-RO-LNN]{morphologically-symmetric reduced-order Lagrangian neural network}
\newacro{pca}[PCA]{principal component analysis}
\newacro{iqr}[IQR]{interquartile range}

\begin{abstract}
Robotic manipulation of deformable objects increasingly relies on learned dynamics models and policies, whose data efficiency and generalization improve markedly by embedding known physical structure as inductive bias. 
This paper studies \emph{morphological symmetries} as one such structure, inherent to many deformable objects yet largely ignored, that arises from dynamically-equivalent body parts such as the two ends of a rope, or the sleeves of a shirt.
We formalize them as relabelings of material points combined with rotations and reflections in $\orthogonalgroup{3}$, and derive the conditions under which the dynamics of continuous and discretized deformable objects are equivariant.
We then establish that the morphological symmetry group exploitable in a manipulation problem is the intersection of the symmetry groups of all its constituents, i.e., shape, mass, material, environment, task, and discretization, and introduce a lattice-based taxonomy to identify it in practice.
Our experiments show that simulators and learned dynamics models do not preserve these symmetries by default, and that policies exploiting them generalize to unseen symmetric scenes.
\end{abstract}

\vspace{-0.2cm}
\section{Introduction}
\label{sec:introduction}

Manipulation of deformable objects, such as garments, cables, and soft tissues remains an open challenge in robotics~\cite{yin2021modeling}.
Precise manipulation requires models that predict the motions of a highly-underactuated object under the forces applied by a robot and, conversely, the forces that steer it towards a desired goal. 
Although deformable objects adhere to well-established laws of continuum mechanics~\cite{marsden1983mathematical-553}, practical models require discretizations with hundreds to thousands of \acp{dof}, and their material properties are generally unknown and difficult to identify~\cite{yin2021modeling}. 
Learned dynamics models and policies circumvent manual derivations, but typically require large amounts of data that are often scarce and expensive in robotics, generalize poorly beyond their training distribution, and lack interpretability.
Embedding prior physical knowledge of the system as inductive bias has been shown to mitigate these limitations~\cite{Karniadakis21:PINN, Lutter2023DeLaN, friedl2024riemannian-dc3, apraez2025morphological}.

\begin{figure}[t]
    \centering
    \adjustbox{trim=0cm 0.cm 0cm 0cm}{\includesvg[width=\linewidth]{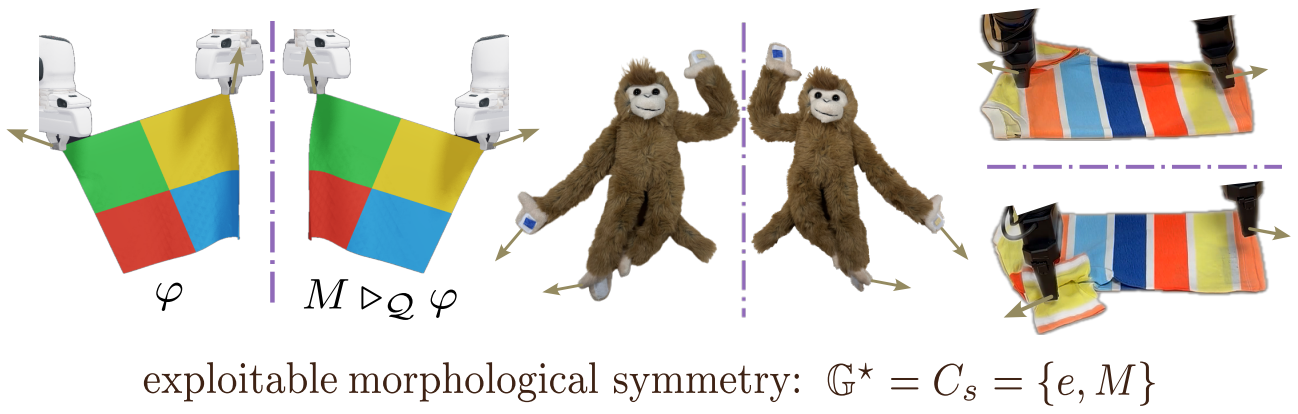}}
    \caption{Morphological symmetries of manipulating deformable objects. Mirroring and relabeling a configuration $\varphi$ yields another physically-admissible configuration $M\lgroupactionon{\configmanifold}{\varphi}$. 
    }
    \label{fig:symmetricsystems}
    \vspace{-0.4cm}
\end{figure}

In addition to their governing mechanical laws, many deformable objects possess a relevant, yet largely unexploited structural prior: \emph{morphological symmetries}.
They arise from dynamically-equivalent body parts~\cite{apraez2025morphological}, e.g., the two ends of a rope, the left and right halves of a shirt, or the rotational symmetry of a fruit.
Morphological symmetries relabel the material points of the object, so that each point takes the role of its symmetric partner, and rotate or reflect the result (Fig.~\ref{fig:symmetricsystems}).
They can be identified from basic structural knowledge of the system and, enforced in learned models, guarantee the transfer of learned behaviors to all symmetric situations, including those never seen during training.

Leveraging morphological symmetries first requires identifying the exploitable symmetry group of a manipulation problem, as it determines symmetric situations. 
This group is not determined by the shape of the object alone, but by all constituents of the problem, i.e., the object (shape, mass distribution, and material), the environment (gravity and contacts), the task, and the discretization used for perception or simulation. 
Exploiting too small a group discards relevant structure, while exploiting too large a one imposes constraints that the data violate.
While morphological symmetries are well understood for rigid-body systems~\cite{apraez2025morphological}, there is currently no systematic way to identify, classify, and incorporate those of deformable objects in robot learning.

In this paper, we close this gap and contribute: 
\emph{(i)} A formalization of the morphological symmetries of deformable objects as relabelings of material points combined with rotations or reflections in $\orthogonalgroup{3}$, together with the conditions under which the continuous and discretized dynamics are equivariant; 
\emph{(ii)} An intersection principle establishing the symmetry group exploitable in a manipulation problem as the intersection of the symmetry groups of all its constituents; and \emph{(iii)} A lattice-based taxonomy of the candidate groups, populated with common deformable objects, to identify this group in practice.
Experiments in simulation and on a real robot show that existing simulators and learned dynamics models often break these symmetries, while policies exploiting them generalize to unseen symmetric scenes. Videos are available at \url{https://sites.google.com/view/morphsym-deformables}.

\section{Related Work}
\label{sec:related_work}

\textbf{Learning-based deformable object manipulation.}
Learned dynamics models of deformable objects typically represent the object as mesh nodes~\cite{pfaff2021learning}, particles~\cite{shi2024robocraft}, or hybrid particle-grid representations~\cite{zhang2024particle} processed by \acp{gnn}, and are combined with \ac{mpc} for manipulation~\cite{zhang2024particle, ma2022learning}.
Their inductive biases (locality, permutation equivariance over nodes, and translation invariance) are generic, and physical knowledge enters, if at all, through material parameters~\cite{zhang2024adaptigraph}.
Alternatively, some policies are learned end-to-end via reinforcement~\cite{wu2019learning} or imitation learning~\cite{black2025pizero}, typically at the cost of even more data.
Both often rely on physics-based simulators for training data or planning, which either exhibit a considerable gap to real deformable objects~\cite{blancomulero2024benchmarking} or require identifying material parameters, e.g., from videos~\cite{jiang2025phystwin}.
None of these approaches exploits the symmetries of the manipulated object, and their meshes or particle samplings may even break them.

\textbf{Physically-structured models.}
Embedding governing equations and conservation laws into learned dynamics models improves generalization and physical plausibility~\cite{Karniadakis21:PINN}.
However, when imposed softly through loss terms, these priors are not guaranteed to hold beyond the training data.
\acp{lnn}~\cite{Lutter2023DeLaN} and \acp{hnn}~\cite{Greydanus2019HNN} instead embed the structure of the equations of motion by construction, thereby enabling structured, e.g., energy-based, control. Both scale to deformable objects via structure-preserving \ac{mor}~\cite{friedl2024riemannian-dc3,friedl2026learning-a88}.
These models exploit the geometry of the configuration space and of the governing equations, but not the symmetries of the modeled object.

\textbf{Symmetries in learning.}
Known symmetries can be incorporated into learned models either through data augmentation~\cite{laskin2020RLdataaug}, which imposes them only approximately, or through equivariant architectures~\cite{weiler2026equivariant}, which enforce them by construction.
Embedding known symmetries improves the accuracy of learned dynamics models~\cite{finzi21:EMLP} and the sample efficiency of manipulation policies~\cite{wang2025practical}. 
However, these works are limited to Euclidean symmetries, which move the object through the workspace as a whole.
Such symmetries are agnostic to the structure of the object, and their reflections yield physically-infeasible configurations.

\textbf{Morphological symmetries.}
Morphological symmetries were formalized for rigid multibody systems with replicated kinematic branches, e.g., humanoid and legged robots, and exploited for dynamics, contact, and policy learning~\cite{apraez2025morphological}.
We formalize them for continuum bodies, where the permutation of limbs becomes a relabeling of material points.
Our intersection principle and taxonomy build on relabeling symmetries in continuum mechanics, which are associated with conservation laws and constrained by material symmetries~\cite{marsden1983mathematical-553}, and on the point groups of molecules in chemistry~\cite{Cotton1990}.

\section{Background}
\label{sec:background}

\subsection{Modeling Deformable Objects}
\label{subsec:def_obj_background}

We model a deformable object as a continuum body~\cite{marsden1983mathematical-553} with reference rest configuration $\body \!\subset\! \euclideanspace^3$ and material points $\bm X \!\in\! \body$.
This description covers volumetric objects as well as linear and planar ones, which are commonly reduced to a curve along their centerline or a surface through their mid-plane.
A configuration is a smooth embedding ${\varphi : \body \to \euclideanspace^3}$ assigning a position $\bm x \!=\! \varphi(\bm X)$ to each material point. Its deformation gradient ${\bm F \!=\! \nabla_{\bm X} \varphi \in \euclideanspace^{3 \times 3}}$ describes the local deformation. Admissible configurations preserve orientation, i.e., $\det \nabla_{\bm X}\varphi \!>\! 0$, and form the infinite-dimensional configuration manifold $\configmanifold \!=\! \{ \varphi \in\mathrm{Emb}^\infty(\body, \euclideanspace^3): \det \nabla_{\bm X}\varphi > 0 \}$. Its tangent space $\configtangentspace{\varphi}$ consists of velocity fields ${\dot\varphi \!:\! \body \!\to\! \euclideanspace^3}$, and its cotangent space $\configcotangentspace{\varphi}$ of force densities ${b \!:\! \body \!\to\! \euclideanspace^3}$, which act on velocity fields through the virtual power ${\langle b, \dot\varphi \rangle \!=\! \int_{\body} b \cdot \dot\varphi \, \mathrm d\bm X}$. The dynamics of the body follow from the Lagrangian $\mathcal L(\varphi, \dot\varphi) \!=\! T(\dot\varphi) - V_{\mathrm{gr}}(\varphi) - V_{\mathrm{el}}(\varphi)$, with kinetic, gravitational, and elastic energies
\begin{equation}
\begin{aligned}
\label{eq:energies}
    T(\dot\varphi) = \frac12 \int_{\body} \rho \, |\dot\varphi|^2 \, \mathrm d\bm X,& \quad
    V_{\mathrm{gr}}(\varphi) = - \int_{\body} \rho \, \gvec \cdot \varphi \, \mathrm d\bm X, \\
    V_{\mathrm{el}}(\varphi) =& \int_{\body} W(\bm X, \nabla_{\bm X}\varphi) \, \mathrm d\bm X, 
\end{aligned}
\end{equation}
where $\rho : \body \to \euclideanspace_{>0}$ is the mass density, $\gvec$ the gravity vector, and $W(\bm X, \bm F)$ the elastic energy density, assumed independent of the deformation history as standard in hyperelasticity.
The associated forced Euler--Lagrange equation is the \ac{pde} of elastodynamics
\begin{equation}
\label{eq:elastodynamics}
    \rho\ddot\varphi = \nabla_{\bm X} \cdot \bm P + \rho \gvec + b,
    \quad
    \bm P(\bm F) = \frac{\partial W(\bm X,\bm F)}{\partial \bm F},
\end{equation}
$\forall \bm X \!\in\! \body$, with $\bm P$ the first Piola--Kirchhoff tensor, i.e., the elastic force per unit rest area,  $\nabla_{\bm X} \cdot$ the divergence and $b$ the density of external forces. Dissipative effects, e.g., friction or air resistance, are included as state-dependent forces $b(\varphi, \dot\varphi)$.

The environment acts on the object through its boundary $\partial\body = \partial\body_{\mathrm N} \cup \partial\body_{\mathrm D}$, either by applying forces or by imposing positions. 
On the Neumann boundary $\partial\body_{\mathrm N}$, it applies a traction ${\bm{t} \!:\! \partial\body_{\mathrm N} \!\to\! \euclideanspace^3}$, i.e., a force per unit rest area, e.g., through a pushing finger or a supporting table,
which the internal stress has to balance as $\bm P \bm n = \bm{t}$, with $\bm n$ the outward unit normal.
On the Dirichlet boundary $\partial\body_{\mathrm D}$, it imposes the positions, $\varphi = \varphi_{\mathrm D}$, e.g., on the part held by a gripper.

In practice, the configuration is approximated by the positions $\bm x_i = \varphi(\bm X_i)$ of $N$ nodes $\{\bm X_i\}_{i=1}^N \subset \body$, e.g., mesh vertices, particles, or tracked points, stacked into ${\q \!\in\! \redmanifold \!=\! \euclideanspace^{3N}}$, the discretized configuration space. The Lagrangian becomes $\mathcal L(\q, \dot\q) = \frac12 \dot\q^\trsp \bm M(\q) \dot\q - V(\q)$, with the symmetric positive-definite mass matrix $\bm M(\q) \in \spd{3N}$ assembled from $\rho$, and the potential $V$ from the elastic and gravitational energies. It yields the standard equations of motion
\begin{equation}
\label{eq:eom}
    \bm M(\q) \ddot\q + \bm C(\q, \dot\q)\dot\q + \bm g(\q) = \bm\tau,
\end{equation}
with Coriolis matrix $\bm C$, conservative forces $\bm g \!=\! \frac{\partial V}{ \partial \q}$, and generalized forces $\bm\tau$. The forward dynamics are $\ddot\q \!=\! f(\q, \dot\q, \bm\tau)$.

\subsection{Groups and Symmetries}
\label{subsec:groups}
Symmetries are invertible transformations that preserve some properties of a system. They form a group, which we formalize in the notation of~\cite[App.~B]{weiler2026equivariant}. 
A group $\group$ is a set equipped with an associative operation $\circ : \group \times \group \to \group$, an identity element $e \in \group$, and an inverse $g^{-1} \in \group$ for each element $g \in \group$. 
A subset $\mathbb H \subseteq \group$ that is itself a group under $\circ$ is a subgroup, denoted $\mathbb H \le \group$.
A group transforms a mathematical object through its action.
A left action of $\group$ on a set $\mathcal X$ is a map $\lgroupactiononblank{\mathcal X} : \group \times \mathcal X \to \mathcal X$ that satisfies $(g_1 \circ g_2) \lgroupactionon{\mathcal X}{x} = g_1 \lgroupactionon{\mathcal X}{(g_2 \lgroupactionon{\mathcal X}{x})}$ and $e \lgroupactionon{\mathcal X}{x} = x$.
A right action $\rgroupactiononblank{\mathcal X}: \mathcal X \times \group \to \mathcal X$ is defined analogously.
A linear action on a vector space $\mathcal X$ is given by a representation $\psi_{\mathcal X} \!:\! \group \!\to\! \generallineargroup{\mathcal X}$, i.e., $g \lgroupactionon{\mathcal X}{x} = \psi_{\mathcal X}(g)\, x$ with $\psi_{\mathcal X}(g_1 \circ g_2) = \psi_{\mathcal X}(g_1)\, \psi_{\mathcal X}(g_2)$, where $\generallineargroup{\mathcal X}$ is the group of invertible matrices acting on $\mathcal X$.\looseness-1

A map $f : \mathcal X \to \mathcal Y$ is $\group$-invariant if $f(g \lgroupactionon{\mathcal X}{x}) = f(x)$ for all $g \in \group$ and $x \in \mathcal X$, i.e., transforming the input does not affect the output. It is $\group$-equivariant if $f(g \lgroupactionon{\mathcal X}{x}) = g \lgroupactionon{\mathcal Y}{f(x)}$, i.e., the output transforms along with the input.

We consider rotations and reflections about the center of the rest configuration $\body$ placed at the origin. They form the orthogonal group $\orthogonalgroup{3} = \{\bm Q \in \euclideanspace^{3 \times 3} \!:\! \bm Q^\trsp \bm Q = \mathbf I \}$, whose elements preserve lengths and angles and either preserve or reverse orientation ($\det \bm Q = \pm 1$). 
Every symmetry group in this paper is a subgroup $\group \le \orthogonalgroup{3}$, whose representations $\rep{\euclideanspace^3}(g) \!=\! \bm Q$ on $\euclideanspace^3$ are orthogonal matrices. 
The same element $g$ acts differently on different objects, e.g., it acts on a set $\set \subset \euclideanspace^3$ and on a scalar field $\phi: \euclideanspace^3 \to \euclideanspace$ as
\begin{equation*}
\label{eq:act-set}
    g \lgroupactionon{\set}{\set} = \{g \lgroupactionon{\euclideanspace^3}{\bm X}: \bm X \in \set\}, 
    \text{ and }
    (\rgroupactionon{}{\phi} \, g)(\bm X) = \phi\big(g \lgroupactionon{\euclideanspace^3}{\bm X}\big).
\end{equation*}
Throughout the paper, groups act from the right on functions of material points, such as $\varphi$, $\rho$, and $W$, by evaluating them at the transformed point as for $\phi$ above.
The associated left action is the right action of $g^{-1}$, i.e., $(g \lgroupactionblank \phi)(\bm X) \!=\! \phi(g^{-1} \lgroupactionblank{\bm X})$.

The \emph{symmetry group}, or \emph{stabilizer}, $\group_x$ of an object $x$ is the set of transformations that leave it unchanged,
\begin{equation}
\label{eq:stabilizer}
    \group_x = \{\, g \in \orthogonalgroup{3} \;:\; g \lgroupactionon{\mathcal X}{x} = x \,\}.
\end{equation}
It is a subgroup of $\orthogonalgroup{3}$, since $g \circ h$ and $g^{-1}$ leave $x$ unchanged whenever $g$ and $h$ do. It is also denoted as $\operatorname{Stab}_{x}$ and referred to as the isotropy group in the literature. For functions of material points, the stabilizer can equivalently be stated with the right action, as $\rgroupactionon{}{x}\, g = x$ iff $g^{-1} \lgroupactionblank x = x$, i.e., $g^{-1} \in \group_x$ and thus $g \in \group_x$.
\section{Morphological Symmetries and the Intersection Principle}
\label{sec:morphological_symmetry}
This section formalizes morphological symmetries of deformable objects and establishes the intersection principle to identify the largest exploitable symmetry group.

\subsection{Morphological Actions}
\label{subsec:morph-actions}
Any symmetry of a deformable object must first be a symmetry of its shape, i.e., map its rest configuration $\body$ onto itself.
Such transformations form the \emph{shape stabilizer}
\begin{equation}
\label{eq:Gshape}
    \Gshape = \{ g \in \orthogonalgroup{3} : g \lgroupactionon{\set}{\body} = \body \}.
\end{equation}
For instance, the shape stabilizer of a square cloth (see Fig.~\ref{fig:cloth-flow}) is $\Gshape = D_{4h}$, the dihedral group of rotations, reflections, and face-flips of a square.
Each $g \! \in \! \Gshape$ assigns a symmetric partner $g \lgroupactionon{\euclideanspace^3}{\bm X} \!\in\! \body$ to every material point $\bm X$, thereby defining the right action
$    (\rgroupactionon{\configmanifold}{\varphi} \, g^{-1})(\bm X) = \varphi\big(g^{-1} \lgroupactionon{\euclideanspace^3}{\bm X}\big) $,
which \emph{relabels} the material points.
The \emph{morphological action} combines this relabeling with the corresponding spatial transformation of $\varphi(\bm{X})\in\euclideanspace^3$ as
\begin{equation}
\label{eq:act-cont}
    (g \lgroupactionon{\configmanifold}{\varphi})(\bm X) 
    = g \lgroupactionon{\euclideanspace^3}{\varphi\big(g^{-1} \lgroupactionon{\euclideanspace^3}{\bm X} \big)}.
\end{equation}
The inner $g^{-1}$ relabels the material point $\bm X$ to its symmetric partner, the outer $g$ rotates and/or reflects the position of this partner.
Together, they form a left action on $\configmanifold$ that leaves the rest configuration unchanged. 
The action~\eqref{eq:act-cont} induces actions on velocity fields, force fields, and deformation gradients.
As it is linear in $\varphi$ and $\rep{\euclideanspace^3}(g)$ is orthogonal, its tangent and cotangent lifts act identically on velocity and force fields  
while deformation gradients transform by conjugation, i.e.,
\begin{align}
\label{eq:act-tang}
    (g \lgroupactionon{\configtangentspace{\varphi}}{\dot\varphi} )(\bm X) &= g \lgroupactionon{\euclideanspace^3}{\dot\varphi\big(g^{-1} \lgroupactionon{\euclideanspace^3}{\bm X}\big)}, \\
\label{eq:act-cotang}
    (g \lgroupactionon{\configcotangentspace{\varphi}}{b})(\bm X) &= g \lgroupactionon{\euclideanspace^3}{b\big(g^{-1} \lgroupactionon{\euclideanspace^3}{\bm X}\big)}, \\
\label{eq:act-F}
    (g \lgroupactionblank\bm F)(\bm X) &=
    \rep{\euclideanspace^3}(g) \, \bm F\big(g^{-1} \lgroupactionon{\euclideanspace^3}{\bm X}\big) \, \rep{\euclideanspace^3}^{\trsp}(g),
\end{align}
where~\eqref{eq:act-F} follows from~\eqref{eq:act-cont} by the chain rule. 
Moreover, the morphological action preserves the virtual power $\langle b, \dot\varphi \rangle$.

\definecolor{nfBlue}{HTML}{A2C9E8}   \definecolor{nfTan}{HTML}{E6C89D}    \definecolor{nfInk}{HTML}{494949}
\definecolor{nfPurple}{HTML}{A791C3} \definecolor{nfTaupe}{HTML}{A58B7E}  \definecolor{nfPanel}{HTML}{F1EDE9}
\definecolor{nfNavy}{HTML}{46598A}   \definecolor{nfMagenta}{HTML}{C68EA7}\definecolor{nfOrange}{HTML}{CA7C11}
\definecolor{nfTeal}{HTML}{77B19A}   \definecolor{nfSlate}{HTML}{5F75A1}  \definecolor{nfGreen}{HTML}{144F44}
\definecolor{nfBrown}{HTML}{472610}
\colorlet{cflcloth}{nfBlue}
\colorlet{cflink}{nfInk}
\colorlet{cflwood}{nfTan}   
\colorlet{cflCshape}{nfNavy}   \colorlet{cflCmass}{nfMagenta} \colorlet{cflCmat}{nfOrange}
\colorlet{cflCenv}{nfTeal}     \colorlet{cflCtask}{nfTaupe}   \colorlet{cflCdisc}{nfSlate}
\colorlet{cflCmorph}{nfGreen}  \colorlet{cflCstar}{nfBrown}
\tikzset{
  cflln/.style={draw=cflink, line width=0.4pt, line join=round, line cap=round},
  cflfine/.style={draw=cflink!55, line width=0.25pt, line cap=round},
  cflmir/.style={draw=nfPurple, dash pattern=on 2.4pt off 0.7pt on 0.5pt off 0.7pt, line width=0.5pt},
  cflarr/.style={-{Stealth[length=4pt, width=3.5pt]}, line width=0.6pt},
  cfljoin/.style={circle, draw=nfInk, fill=white, line width=0.6pt, inner sep=1pt, font=\scriptsize},
  cfltag/.style={font=\scriptsize\sffamily\bfseries, inner sep=1.5pt, anchor=south},
  cfllevel/.style={font=\scriptsize, text=nfInk, rotate=90, inner sep=0pt},
  cflshape/.style={draw=cflCshape}, cflmass/.style={draw=cflCmass}, cflmat/.style={draw=cflCmat},
  cflenv/.style={draw=cflCenv, fill=nfTeal!7}, cfltask/.style={draw=cflCtask, fill=nfPanel},
  cfldisc/.style={draw=cflCdisc, fill=nfSlate!8},
  cflmorph/.style={draw=cflCmorph, line width=1pt, fill=nfGreen!7},
  cflstar/.style={draw=cflCstar, line width=1pt, fill=nfPanel},
}
\providecommand{\cflMirror}[1]{%
  \ifcase#1\or \draw[cflmir] (0,-0.56) -- (0,0.56);%
  \or \draw[cflmir] (-0.56,0) -- (0.56,0);%
  \or \draw[cflmir] (-0.52,-0.52) -- (0.52,0.52);%
  \or \draw[cflmir] (-0.52,0.52) -- (0.52,-0.52);%
  \fi}
\providecommand{\cflFlip}{\node[font=\scriptsize\sffamily\bfseries, fill=nfPurple, text=white, circle, inner sep=0.8pt]} 
\ifdefined\cflFbox\else\newsavebox\cflFbox\fi
\sbox\cflFbox{\scriptsize\sffamily\bfseries F} 
\providecommand{\cflMz}{\makebox[\wd\cflFbox][c]{\raisebox{0pt}[\ht\cflFbox][0pt]{\resizebox{!}{\ht\cflFbox}{\raisebox{\depth}{$\bm{\updownarrow}$}}}}}
\DeclareRobustCommand{\cflSqIcon}{\tikz[baseline=-0.6ex]\draw[cflln, fill=cflcloth] (-0.085,-0.085) rectangle (0.085,0.085);}
\providecommand{\cflLegFont}{\scriptsize}
\providecommand{\cflGravSym}{\node[circle, draw=cflink, fill=white, line width=0.45pt, inner sep=0pt, minimum size=7.2pt, anchor=center,
  path picture={\draw[cflink, line width=0.45pt] (path picture bounding box.south west) -- (path picture bounding box.north east)
    (path picture bounding box.north west) -- (path picture bounding box.south east);}]}
\providecommand{\cflCore}[4]{%
    \draw[rounded corners=1.5pt, fill=white, #1] (-0.7,-0.7) rectangle (0.7,0.7);
    #2
    \foreach \m in {#3} {\cflMirror{\m}}
    \ifnum#4=1 \cflFlip at (0.44,0.44) {\cflMz};\fi 
    \ifnum#4=2 \cflGravSym at (0.44,0.44) {};\fi} 
\providecommand{\cflPic}[8]{%
  \begin{scope}[shift={(#1,#2)}, scale=0.6]
    \cflCore{cfl#3}{#4}{#5}{#6}
    \node[font=\scriptsize, anchor=north, inner sep=1pt, align=center, text=cflC#3] at (0,-0.74) {#7\\[-1pt]#8};
  \end{scope}}
\providecommand{\cflPicSide}[8]{%
  \begin{scope}[shift={(#1,#2)}, scale=0.6]
    \cflCore{cfl#3}{#4}{#5}{#6}
    \node[font=\scriptsize, anchor=west, inner sep=2pt, align=left, text=cflC#3] at (0.72,0) {#7\\[-1pt]#8};
  \end{scope}}
\providecommand{\cflSq}{\draw[cflln, fill=cflcloth] (-0.42,-0.42) rectangle (0.42,0.42);}
\providecommand{\cflWeave}{\foreach \t in {-0.28,-0.14,0,0.14,0.28} {\draw[cflfine] (-0.42,\t) -- (0.42,\t); \draw[cflfine, draw=cflink!25] (\t,-0.42) -- (\t,0.42);}}
\providecommand{\cflHem}{\draw[nfMagenta, line width=1.5pt] (-0.42,0.39) -- (0.42,0.39);}
\providecommand{\cflTable}{\fill[cflwood] (-0.6,-0.6) rectangle (0.6,0.6);}
\providecommand{\cflMeshDots}{\foreach \i in {-0.42,-0.14,0.14,0.42} {\foreach \j in {-0.42,-0.14,0.14,0.42} \fill[cflink] (\i,\j) circle[radius=0.045];}}
\providecommand{\cflTarget}{\draw[nfTaupe, densely dotted, line width=0.6pt] (-0.5,-0.5) rectangle (0.5,0.5);}
\providecommand{\cflMesh}{\foreach \i in {-0.42,-0.14,0.14,0.42} {\draw[cflfine] (\i,-0.42) -- (\i,0.42); \draw[cflfine] (-0.42,\i) -- (0.42,\i);
    \foreach \j in {-0.42,-0.14,0.14,0.42} \fill[cflink] (\i,\j) circle[radius=0.045];}}
    
\begin{figure}[t]
\centering
\begin{tikzpicture}[x=1cm, y=1cm]
  \draw[-{Stealth[length=3pt, width=3pt]}, nfInk!80, line width=0.4pt] (-0.28,0.42) -- (-0.28,-4.3)
    node[midway, rotate=90, anchor=south, font=\scriptsize, inner sep=1pt, text=nfInk] {fewer symmetries};
  \draw[cflCshape, line width=0.8pt] (0.7,-0.42) -- (1.3,-1.0);
  \draw[cflCshape, line width=0.8pt] (0.4,-2.58) -- (0.43,-0.42) node[pos=0.42, sloped, above, font=\tiny\itshape, inner sep=1.5pt] {subgroup};
  \cflPicSide{0.5}{0}{shape}{\cflSq}{1,2,3,4}{1}{shape}{$\Gshape{=}D_{4h}$}
  \cflPic{5.0}{0}{task}{\cflSq\cflTarget}{1,2,3,4}{1}{task}{$\Gtask{=}D_{4h}$}
  \cflPic{6.8}{0}{disc}{\cflSq\cflMesh}{1,2,3,4}{1}{discretization}{$\Gdisc{=}D_{4h}$}
  \cflPic{1.45}{-1.42}{mat}{\cflSq\cflWeave}{1,2}{1}{material}{$\Gmat{=}D_{2h}$}
  \cflPic{3.4}{-1.42}{env}{\cflTable\cflSq}{1,2,3,4}{2}{environment}{$\Genv{=}C_{4v}$}
  \cflPic{0.5}{-3.0}{mass}{\cflSq\cflHem}{1}{1}{mass}{$\Gmass{=}C_{2v}$}
  \node[cfljoin] (j1) at (1.45,-3.0) {$\cap$};
  \cflPic{2.45}{-3.0}{morph}{\cflSq\cflWeave\cflHem}{1}{1}{\textbf{morphological}}{$\Gmorph{=}C_{2v}$}
  \draw[cflarr, cflCmass] (0.92,-3.0) -- (j1);
  \draw[cflarr, cflCmat] (1.45,-2.5) -- (j1);
  \draw[cflarr, cflCmorph] (j1) -- (2.01,-3.0);
  \node[cfljoin] (j2) at (4.3,-3.0) {$\cap$};
  \draw[cflarr, cflCmorph] (2.89,-3.0) -- (j2);
  \draw[cflarr, cflCenv] (3.6,-2.5) -- (j2);
  \draw[cflarr, cflCtask] (5.0,-1.1) -- (j2);
  \draw[cflarr, cflCdisc] (6.8,-1.1) -- (j2);
  \cflPicSide{4.3}{-3.9}{star}{\cflTable\cflSq\cflWeave\cflHem\cflMeshDots}{1}{2}{\textbf{exploitable}}{$\Gstar{=}C_s{=}\{e, \Mx\}$}
  \draw[cflarr, cflCstar] (j2) -- (4.3,-3.46);
  \node[font=\cflLegFont, inner sep=0pt, anchor=north] at (3.63,-4.42) {%
    \tikz[baseline=-0.5ex]\draw[cflmir] (0,0) -- (3.2mm,0);\ mirror plane \enspace
    \tikz[baseline=(f.base)]\cflFlip (f) {\cflMz};\ mirror $M_z$ \enspace
    \tikz[baseline=-0.6ex]\cflGravSym {};\ gravity $\gvec$ \enspace
    \tikz[baseline=(n.base)]\node[cfljoin, inner sep=0.5pt, font=\cflLegFont] (n) {$\cap$};\ intersection \enspace
    \tikz[baseline=-0.5ex]\fill[cflwood] (0,-0.07) rectangle (0.22,0.07);\ table};
\end{tikzpicture}
\caption{Intersection principle for a hemmed square cloth~(\cflSqIcon) with an unbalanced weave on a table. Rows share the same group order.}
\label{fig:cloth-flow}
\vspace{-0.5cm}
\end{figure}
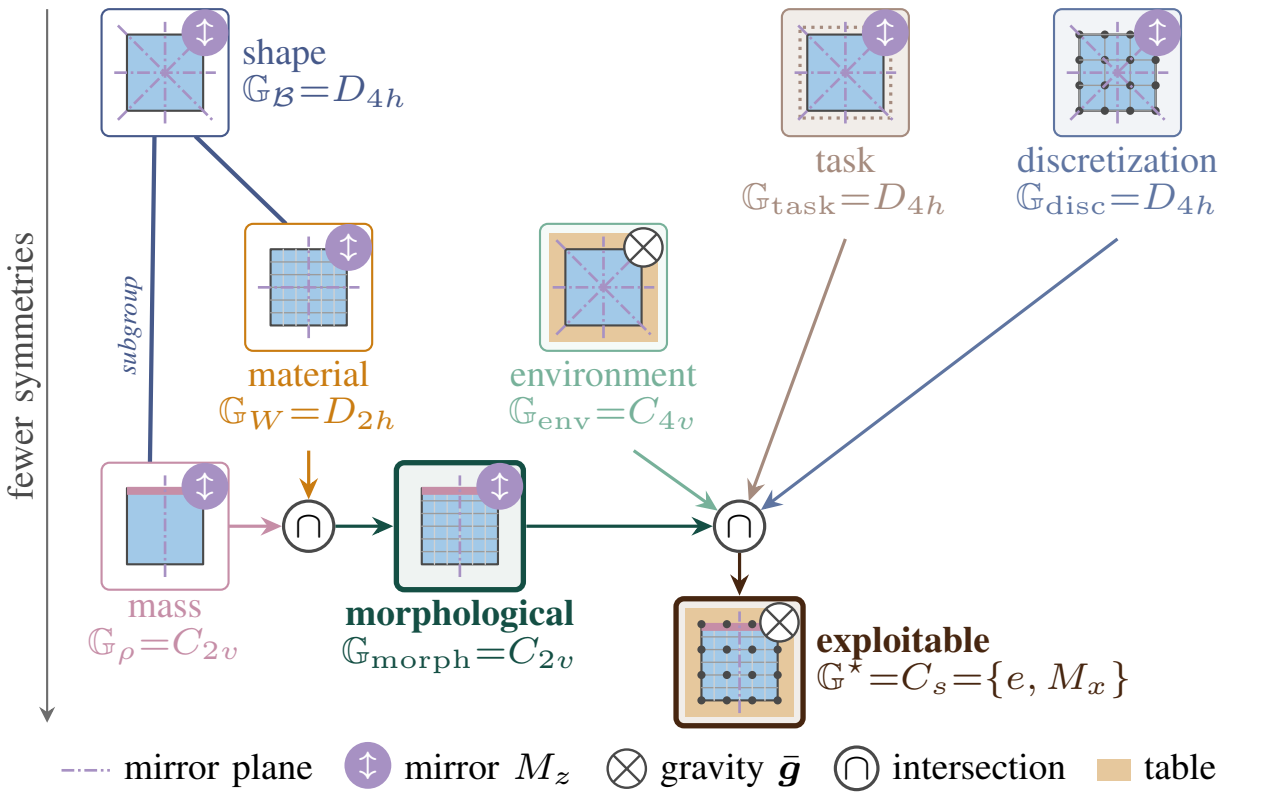

The morphological action~\eqref{eq:act-cont} differs from the commonly-used Euclidean action $(g \lgroupactionon{\euclideanspace^3}{\varphi})(\bm X) = g \lgroupactionon{\euclideanspace^3}{\varphi(\bm X)}$, which rotates or reflects the object as a whole without relabeling, in two respects.
First, Euclidean symmetries apply to every object, while morphological symmetries require symmetric shapes. 
Second, as illustrated in Fig.~\ref{fig:symmetricdynamics}-\emph{left}, a Euclidean reflection yields an inverted, physically-infeasible configuration, whereas action~\eqref{eq:act-cont} preserves orientation, 
yielding admissible configurations. This is the deformable counterpart of the feasible symmetries of~\cite{apraez2025morphological}, where a robot emulates an infeasible reflection by permuting its limbs.

Both morphological and Euclidean actions are special cases of a general construction, which transforms the domain and codomain of a configuration $\varphi \!:\! \body \!\to\! \euclideanspace^3$ separately. Its domain, i.e., material side, admits a relabeling $h \in \Gshape$, and its codomain, i.e., spatial side, an isometry $u\in\euclideangroup{3}$, $u \lgroupactionon{\euclideanspace^3}{\bm x} = \bm Q_u \bm x + \bm d_u$, which moves the object as a whole, as the base motions of floating-base robots~\cite{apraez2025morphological}. Combining them gives the action of a pair ${ (u, h) \lgroupactionon{\configmanifold}{\varphi} = u \circ \varphi \circ h^{-1} }$, which yields an admissible configuration iff ${\det \bm Q_u \!=\! \det h}$. The Euclidean action is the pair $(u, e)$, and the morphological action is the matched pair $(g, g)$. Every admissible pair factors as $(u, h) = (u \circ h^{-1}, e) \circ (h, h)$, i.e., as a morphological action followed by a rigid motion $u \circ h^{-1} \in \speceuclideangroup{3}$. 

\subsection{Morphological Symmetry Group of a Deformable Object}
\label{subsec:morph-group}

\begin{figure*}[t]
    \centering
    \includesvg[width=\linewidth]{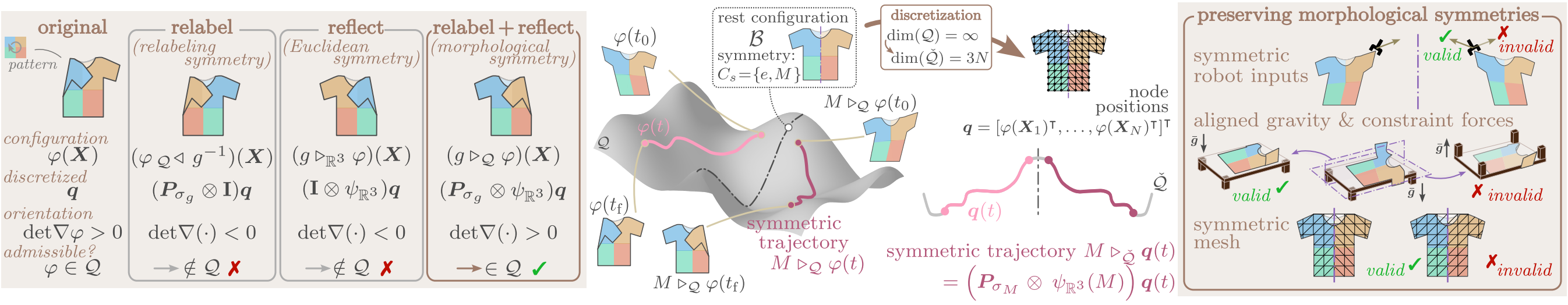}
    \caption{Morphological symmetries of a shirt with a single mirror, i.e., with stabilizer $C_s$. \emph{Left:} Relabeling or reflecting (Euclidean action) alone inverts the material ($\det\nabla(\cdot)<0$), whereas their combination (morphological action) yields admissible configurations.
    \emph{Middle:} Symmetric initial conditions and inputs yield symmetric trajectories on the configuration manifold $\configmanifold$ and its discretization $\redmanifold$.
    \emph{Right:} The symmetry is preserved only under symmetric robot inputs, aligned gravity and constraint forces, and symmetric discretization.}
    \label{fig:symmetricdynamics}
    \vspace{-0.4cm}
\end{figure*}

The morphological action is defined for every shape symmetry. 
However, it is a symmetry of the dynamics, i.e., it maps motions of the object onto physically-valid symmetry-transformed ones (see Fig.~\ref{fig:symmetricdynamics}-\emph{middle}), only when it additionally leaves the energies~\eqref{eq:energies} of the Lagrangian unchanged. The kinetic and gravitational energies depend on the mass density $\rho$, and the elastic energy on the material elastic energy density $W$. As both are defined on $\body$, each $g \in \Gshape$ acts on them from the right by evaluating them at the partner point, i.e., $(\rgroupactionon{}{\rho} \, g)(\bm X) \!=\! \rho\big(g \lgroupactionon{\euclideanspace^3}{\bm X})$ and $(\rgroupactionon{}{W} \, g)(\bm X, \bm F) \!=\! W(g \lgroupactionon{\euclideanspace^3}{\bm X}, g \lgroupactionblank\bm F )$.
For $W$, the deformation gradient is additionally conjugated as in~\eqref{eq:act-F}, i.e., along transformed material directions. The symmetries preserving the object mass and material distributions form the stabilizers
\begin{align}
\label{eq:Gmass}
    \Gmass &= \{ g \in \Gshape : \rgroupactionon{}{\rho} \, g = \rho \}, \\
\label{eq:Gmat}
    \Gmat &= \{ g \in \Gshape : \rgroupactionon{}{W} \, g = W \}.
\end{align}
For the square cloth of Fig.~\ref{fig:cloth-flow}, the hem adds mass along one edge, thus removing symmetries moving this edge and reducing $\Gshape = D_{4h}$ to $\Gmass = C_{2v}$. The unbalanced weave makes the two thread directions nonequivalent, thus removing symmetries that exchange them to leave $\Gmat = D_{2h}$.

Note that $\Gmat$ differs from two classical invariances of $W$~\cite{marsden1983mathematical-553}. \emph{Frame indifference} states that rotating the body in space does not change its elastic energy, i.e., ${W(\bm X, \bm R \bm F) \!=\! W(\bm X, \bm F)}$ $\forall\bm R \in \specorthogonalgroup{3}$, i.e. $W$ is invariant under to Euclidean symmetries $(u,e)$, and holds for every object and material.
The \emph{material symmetry group} ${\mathcal G_{\bm X} \!\le\! \orthogonalgroup{3}}$ contains the transformations $\bm Q$ of the material directions at a single point $\bm X$ that do not change its response, i.e., $W(\bm X, \bm Q \bm F \bm Q^\trsp) \!=\! W(\bm X, \bm F)$ $\forall\bm F$. It is $\orthogonalgroup{3}$ for isotropic materials. 
In contrast, $\Gmat$~\eqref{eq:Gmat} compares the material at a point $\bm X$ with that at its partner $g \lgroupactionon{\euclideanspace^3}{\bm X}$, and is thus a property of the body and not of the material alone. For a homogeneous material, $\mathcal G_{\bm X} = \mathcal G$ $\forall \bm X$, and it reduces to $\Gmat = \Gshape \cap \mathcal G$.

\begin{definition}[Morphological symmetry group]
\label{def:Gmorph}
The morphological symmetry group or stabilizer of a deformable object is\looseness-1
\begin{equation}
    \Gmorph = \Gmass \cap \Gmat \leq \Gshape.
\end{equation}
\end{definition}

$\Gshape$ and $\Gmorph$ capture the symmetries of the object at two levels. 
$\Gshape$ suffices for purely shape-dependent parts of the manipulation pipeline, such as perception or keypoint detection. $\Gmorph$ additionally accounts for mass and material and, as shown next, encodes the symmetries of the kinetic and elastic energies, which gravity further restricts.

\begin{proposition}[Per-term symmetry groups]
\label{prop:perterm}
Under the actions~\eqref{eq:act-cont} and~\eqref{eq:act-tang}, the kinetic, gravitational, and elastic energies~\eqref{eq:energies} are respectively invariant under
\begin{equation}
\label{eq:G_perterm}
    \group_{T} = \Gmass, \quad
    \group_{V_{\mathrm{gr}}} = \Gmass \cap \Ggrav, \quad
    \group_{V_{\mathrm{el}}} = \Gmat, 
\end{equation}
with $\Ggrav = \{ g \in \orthogonalgroup{3} : g \lgroupactionon{\euclideanspace^3}{\gvec} = \gvec \}$ the stabilizer of gravity.
Therefore, the Lagrangian $\mathcal L$ is invariant under
\begin{equation}
\label{eq:G_lagrangian}
    \group_{\mathcal{L}} = \group_{T} \cap \group_{V_{\mathrm{gr}}} \cap \group_{V_{\mathrm{el}}} = \Gmorph \cap \Ggrav,
\end{equation}
forming the hierarchy $\group_{\mathcal L} \le \Gmorph \!=\! \Gmass \cap \Gmat \le \Gshape$.
\end{proposition}

For the square cloth of Fig.~\ref{fig:cloth-flow}, $\Gmorph \!=\! C_{2v}$ consists of the mirror $\Mx$, the flip $\Fl$, and their product. When the cloth lies flat, gravity is normal to it. The mirror $\Mx$, in a vertical plane, leaves $\gvec$ unchanged, whereas the flip $\Fl$ reverses it and is therefore excluded, hence $\group_{\mathcal L} \!=\! \{e, \Mx\}$. 

\begin{proposition}[Equivariant continuous dynamics]
\label{prop:equivariance}
Let ${g \!\in\! \group_{\mathcal L}}$, i.e., leaving $\mathcal L$ invariant, and let $\varphi(t)$ solve the forced Euler--Lagrange equations~\eqref{eq:elastodynamics} with force density $b$, traction $\bm t$ on $\partial\body_{\mathrm N}$, and boundary data $\varphi_{\mathrm D}$ on $\partial\body_{\mathrm D}$. 
Then, $g \lgroupactionon{\configmanifold}{\varphi(t)}$ solves~\eqref{eq:elastodynamics} with force density $g \lgroupactionon{\configcotangentspace{\varphi}}{b}$, traction $g \lgroupactionon{\configcotangentspace{\varphi}}{\bm t}$ on $g \lgroupactionon{\set}{\partial\body_{\mathrm N}}$, and boundary data $g \lgroupactionon{\configmanifold}{\varphi_{\mathrm D}}$ on $g \lgroupactionon{\set}{\partial\body_{\mathrm D}}$.
\end{proposition}

Propositions~\ref{prop:perterm}-\ref{prop:equivariance} extend, with $\mathcal L$ up to a constant, from $(g,g)$ to general pairs $(u, h)$ with $h \!\in\! \Gmorph$ and $\bm Q_u \gvec \!=\! \gvec$. By the factorization of Sec.~\ref{subsec:morph-actions}, such a pair only adds the rigid motion $u \circ h^{-1}$, which preserves the kinetic and, by frame indifference, the elastic energy. For $h = e$, these are the Euclidean symmetries of all objects, associated with conserved momenta by Noether's theorem~\cite{marsden1983mathematical-553}, to which morphological symmetries add those specific to the object.

\subsection{Action on the Discretized Body}
\label{sec:act-disc}

Learned models and simulators operate on discretized configurations $\q \!=\! [\varphi(\bm X_1)^\trsp, \dots, \varphi(\bm X_N)^\trsp]^\trsp \!\in\! \redmanifold = \euclideanspace^{3N}$ of nodes $\{\bm X_i\}_{i=1}^N$ (Sec.~\ref{subsec:def_obj_background}).
The stabilizer of this node set is\looseness-1
\begin{equation}
\label{eq:Gdisc}
    \Gdisc = \group_{\{\bm X_i\}} = \{ g \in \orthogonalgroup{3} : g \lgroupactionon{\euclideanspace^3}{\bm X_i} = \bm X_{\sigma_g(i)} \ \forall i \},
\end{equation}
with $\sigma_g$ the induced permutation of node indices. For mesh-based discretizations, $g$ must also preserve connectivity.

\begin{proposition}[Discrete morphological action]
\label{prop:disc-action}
For ${g \!\in\! \Gshape \cap \Gdisc}$, sampling commutes with the action~\eqref{eq:act-cont}, i.e.,
\begin{equation}
\label{eq:act-disc}
    g \lgroupactionon{\redmanifold}{\q} = \rep{\redmanifold}(g) \, \q,
    \qquad
    \rep{\redmanifold}(g) = \bm P_{\sigma_g} \otimes \rep{\euclideanspace^3}(g) ,
\end{equation}
with permutation matrix $(\bm P_{\sigma_g})_{ji} = \delta_{j \sigma_g(i)}$. Moreover, as $\rep{\redmanifold}$ is orthogonal, $g$ acts on velocities and generalized forces as
\begin{equation}
\label{eq:actdisc-tancotan}
    g \lgroupactionon{\redtangentspace{\q}}{\dot{\q}} = \rep{\redmanifold}(g) \, \dot{\q}, \qquad
    g \lgroupactionon{\redcotangentspace{\q}}{\bm\tau} = \rep{\redmanifold}(g) \, \bm\tau.
\end{equation}
\end{proposition}

$\bm P_{\sigma_g}$ relabels the nodes and $\rep{\euclideanspace^3}(g)$ transforms the attached (co)vectors, mirroring the joint-space representation of robots with replicated kinematic branches~\cite{apraez2025morphological}.
In contrast, the Euclidean action $\mathbf I_N \otimes \rep{\euclideanspace^3}(g)$ involves no relabeling. 
For $g \!\notin\! \Gdisc$, no permutation $\sigma_g$ exists and the morphological action~\eqref{eq:act-cont} has no counterpart on $\redmanifold$.
Provided that partner nodes carry equal mass and material parameters, the discretized Lagrangian inherits the invariance of $\mathcal L$ for $g \in \group_{\mathcal L} \cap \Gdisc$, with~\eqref{eq:act-disc} in place of~\eqref{eq:act-cont}, so that symmetry-transformed trajectories persist on $\redmanifold$ (see Fig.~\ref{fig:symmetricdynamics}-\emph{middle}), as follows. 

\begin{proposition}[Equivariant discretized dynamics]
\label{prop:equivariance-disc}
The discretized Lagrangian is invariant under $g \in \group_{\mathcal{L}} \cap \Gdisc$ iff
\begin{align}
\label{eq:disc-massinvariance}
    \bm M(g \lgroupactionon{\redmanifold}{\q}) &= \rep{\redmanifold}(g) \, \bm M(\q) \, \rep{\redmanifold}^\trsp(g),
    \\
\label{eq:disc-potinvariance}
    V(g \lgroupactionon{\redmanifold}{\q}) &= V(\q).
\end{align}
In this case, the forward dynamics $f$ are equivariant, i.e.,
\begin{equation}
\label{eq:eom-equivariance}
    f\big(g \lgroupactionon{\redmanifold}{\q}, g \lgroupactionon{\redtangentspace{\q}}{\dot\q}, g \lgroupactionon{\redcotangentspace{\q}}{\bm\tau} \big) = g \lgroupactionon{\redtangentspace{\q}}{f(\q, \dot\q, \bm\tau)}.
\end{equation}
\end{proposition}

When $\q$ collects nodal positions, the three coordinates of a node share the same inertia, so that $\bm M \!=\! \hat{\bm M} \otimes \mathbf I_3$, with $\hat{\bm M} \!\in\! \euclideanspace^{N \times N}$ the matrix of nodal masses. 
As $\rep{\euclideanspace^3}(g)$ in~\eqref{eq:act-disc} is orthogonal,~\eqref{eq:disc-massinvariance} reduces to $\bm P_{\sigma_g} \hat{\bm M} \bm P_{\sigma_g}^\trsp = \hat{\bm M}$, i.e., partner nodes must have equal mass. Therefore, the kinetic energy only depends on the relabeling and $\rep{\euclideanspace^3}(g)$ enters through $V$.\looseness-1

\subsection{The Exploitable Symmetry in Manipulation Problems}
\label{subsec:exploitable-group}

The stabilizers $\Gmorph$ and $\Gdisc$ capture the symmetries of the object and of its discretization.
However, a manipulated object never moves in isolation, as it is constrained by its environment and moved for a given task. 
An equivariant model can therefore exploit a morphological symmetry only when it is also a symmetry of the environment and of the task.
We refer to the object, environment, task, and discretization as the \emph{constituents} of the problem, whose stabilizers jointly determine its exploitable symmetries.

\begin{definition}[Exploitable symmetry group]
\label{def:exploitable}
Let $\mathcal C(\varphi) \!\in\! \configcotangentspace{\varphi}$ be the constraint force field that the environment exerts on the body, and $c : \configmanifold \to \euclideanspace$ the task cost. Their stabilizers are
\begin{align}
\label{eq:GC}
    \group_{\mathcal C} &= \{ g \in \Gshape : \mathcal C(g \lgroupactionon{\configmanifold}{\varphi}) = g \lgroupactionon{\configcotangentspace{\varphi}}{\mathcal C(\varphi)} \ \forall \varphi \}, \\
\label{eq:Gc}
    \Gtask &= \group_{c} = \{ g \in \Gshape : c(g \lgroupactionon{\configmanifold}{ \varphi}) = c(\varphi) \ \forall \varphi \}.
\end{align}
With the environment stabilizer $\Genv \!=\! \Ggrav \cap \group_{\mathcal C}$,  
the exploitable symmetry group of a manipulation problem is
\begin{equation}
\label{eq:exploitable}
    \Gstar = \Gmorph \cap \Genv \cap \Gtask \cap \Gdisc .
\end{equation}
\end{definition}
The constraint forces $\mathcal{C}$ enter the dynamics through the boundary conditions of Sec.~\ref{subsec:def_obj_background} as the tractions $\bm t$ on $\partial\body_{\mathrm N}$ and the reactions that maintain $\varphi = \varphi_{\mathrm D}$ on $\partial\body_{\mathrm D}$. By Proposition~\ref{prop:equivariance}, the transformed motion $g \lgroupactionon{\configmanifold}{\varphi}$ is a solution only under the transformed forces~\eqref{eq:act-cotang}, which the environment provides iff $g \in \Genv$, e.g., for symmetric robot inputs, and aligned gravity and constraint forces (see Fig.~\ref{fig:symmetricdynamics}-\emph{right}).
For the cloth of Fig.~\ref{fig:cloth-flow}, the task and discretization are symmetric, but gravity and the table remove the flip $\Fl$ and $\Mx \Fl$ from $\Gmorph$, so that ${\Gstar \!=\! \{e, \Mx\}}$. 
The following two principles characterize $\Gstar$.\looseness-1

\begin{principle}[Intersection principle]
\label{prop:intersection}
$\Gstar$ is the largest subgroup of $\orthogonalgroup{3}$ under which all constituents are invariant.
\end{principle}
Indeed, an element $g$ stabilizes all constituents iff it belongs to the intersection of their stabilizers.

\begin{principle}[Monotonicity and maximality]
Adding a constituent can only reduce the exploitable group, so that $\Gstar \le \Gmorph \le \Gshape$. Moreover, a group $\mathbb{H} \le \orthogonalgroup{3}$ leaves all constituents invariant iff $\mathbb{H} \le \Gstar$. Hence, any $\mathbb{H} < \Gstar$ disregards exploitable symmetries of the problem, and any $\mathbb{H} \not\le \Gstar$ imposes symmetries that it does not have.
\end{principle}

Note that, as the only constituent that is a design choice, the discretization should preserve as many symmetries of the others as possible to maximize $\Gstar$.
The exploitable group $\Gstar$ also governs policy design. Consider a manipulation task modeled as a Markov decision process with transitions given by the dynamics and with reward $-c$. As $\Gstar$ preserves both transitions and reward, the optimal value function is $\Gstar$-invariant and there exists an optimal $\Gstar$-equivariant policy $\pi(g \lgroupactionblank \bm a \!\mid\! g \lgroupactionblank \bm s) = \pi(\bm a \!\mid\! \bm s)$~\cite{pol2020mdp-906}, both of which can be learned with $\Gstar$-augmented data or $\Gstar$-equivariant models.
Next, we provide the taxonomy to identify $\Gstar$ in practice.\looseness-1

\section{A Lattice-Based Taxonomy of Morphological Symmetries of Deformable Objects}
\label{sec:taxonomy}

\begin{figure*}[ht]
    \centering
    \includesvg[width=\textwidth]{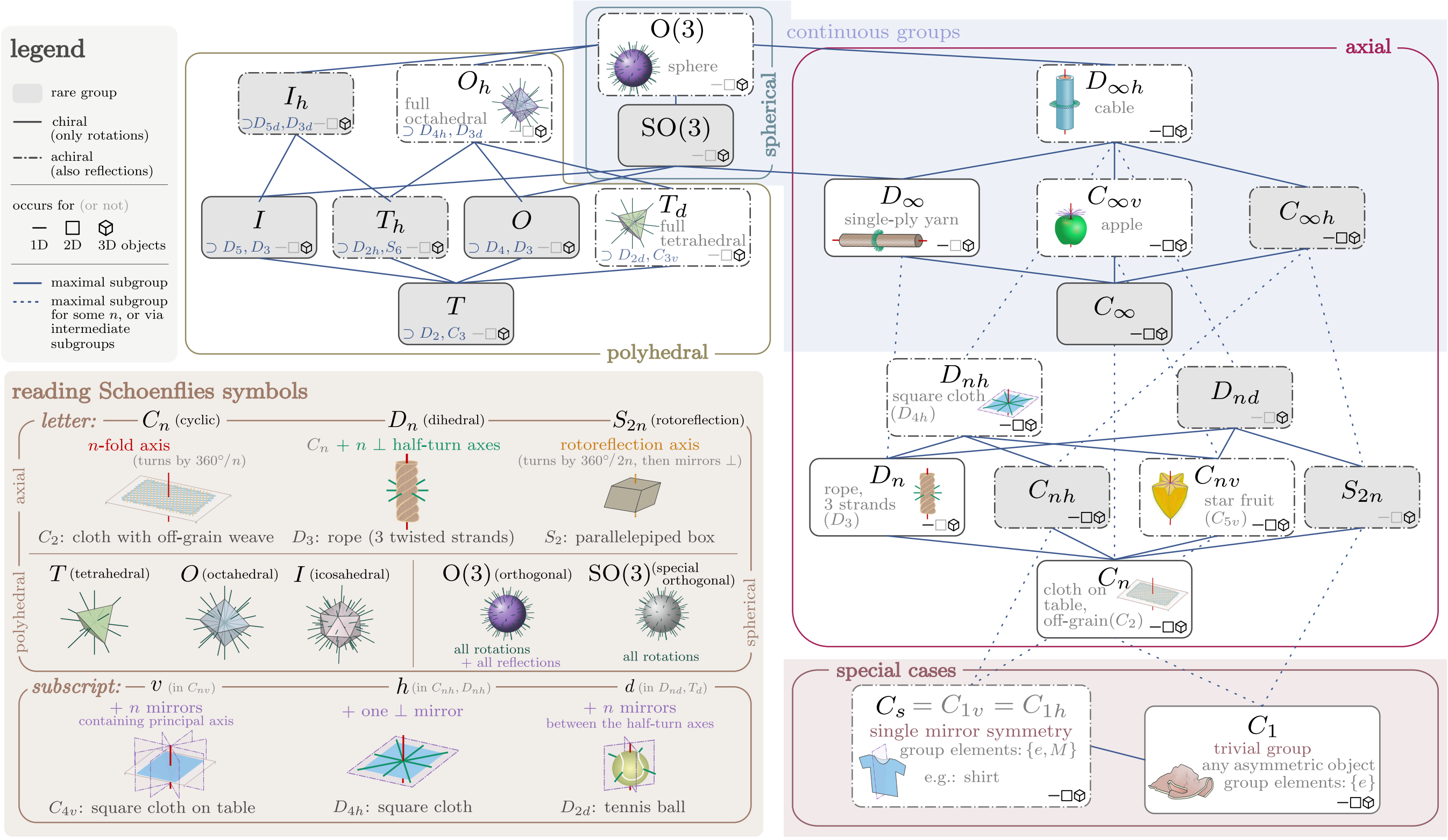}
    \caption{Taxonomy of the morphological symmetry groups of deformable objects with representative examples.
    Edges between the closed subgroups of $\orthogonalgroup{3}$ indicate maximal subgroups, i.e., represent one step of symmetry breaking. \emph{Bottom-left:} Illustration of the reading of a Schoenflies symbol from its letter, order $n$ of its principal axis, and subscript, with examples of objects.}
    \label{fig:grouptaxonomy}
    \vspace{-0.4cm}
\end{figure*}

The intersection principle reduces the identification of $\Gstar$ to \emph{(i)} recognizing the stabilizer of each constituent, and \emph{(ii)} intersecting them. As all stabilizers are closed subgroups of $\orthogonalgroup{3}$, which are completely classified~\cite{olive2019effective-865}, the operations involve, up to the number and orientation of their axes and mirror planes, a set of candidate groups composed of $7$ continuous groups, $7$ polyhedral groups, and $7$ families of axial groups. 
Fig.~\ref{fig:grouptaxonomy} arranges them, with representative deformable objects, in the lattice-based taxonomy described next.\looseness-1

A symmetry group is often named \emph{abstractly}, i.e., by its structure alone, with its action specified separately via a representation. 
Such names can be ambiguous, as a mirror, a half-turn, and a point inversion correspond to the same abstract two-element group $\cyclicgroup{2}$, but represent different transformations that different constituents can break. 
We therefore adopt the Schoenflies notation of molecular symmetry~\cite{Cotton1990}, which names each stabilizer as a \emph{concrete} subgroup of $\orthogonalgroup{3}$. 

Fig.~\ref{fig:grouptaxonomy}-\emph{bottom-left} illustrates the reading of a Schoenflies symbol. The letter indicates the group type. Axial groups have one principal axis. The cyclic group $C_n$ contains the rotations by multiples of $2\pi/n$ about this axis.
The dihedral group $D_n$ adds $n$ half-turns about perpendicular axes, and $S_{2n}$ is generated by a rotoreflection, i.e., a rotation by $\pi/n$ followed by a perpendicular mirror. 
They become the continuous axial groups for $n=\infty$. 
The polyhedral groups $T$, $O$, and $I$ have several axes of order three or higher, and the spherical groups $\specorthogonalgroup{3}$ and $\orthogonalgroup{3}$ contain all rotations, without and with reflections.
Subscripts indicate additional mirrors, namely $h$ for a mirror plane perpendicular to the principal axis, which exchanges and relabels, e.g., two faces of a cloth, $v$ for $n$ planes containing the principal axis, and $d$ for $n$ planes between the half-turn axes.
A single mirror forms the group $C_s\!=\!C_{1v}\!=\!C_{1h}$, the most common case in robotics.\looseness-1

Fig.~\ref{fig:grouptaxonomy} draws the subgroup lattice as a Hasse diagram, where each edge connects a group to a maximal subgroup, so that moving down an edge is one step of minimal symmetry breaking. 
Each node represents a group up to orientation. A stabilizer then is one such group with its axes and planes fixed to the object, e.g., $\{e, \Mx\}$ and $\{e, \Fl\}$ are two different stabilizers on the node $C_s$. 
The dimension of an object already narrows down the candidates. Linear and planar deformable objects, e.g., ropes and cloth, only admit $\Dinfh$ and its subgroups. Among these, the group is determined by the planar symmetry of the cross-section or outline, and by the equivalence of the two ends or faces. Polyhedral and spherical groups only occur for volumetric objects. \looseness-1 

$\Gstar$ is then identified in three steps, illustrated in Fig.~\ref{fig:cloth-flow} for a square cloth. 
First, $\Gshape$ is located in the diagram from the axes and mirrors of $\body$ as above, e.g., $\Gshape = D_{4h}$ for the square cloth.  
Second, each constituent removes the elements of $\Gshape$ that change it, moving down the diagram. The unbalanced weave reduces to $\Gmat \!=\! D_{2h}$, the hem along one edge to $\Gmass \!=\! C_{2v}$ with principal axis in the plane of the cloth. Gravity and the table then only keep $\Genv = C_{4v}$. 
Third, the stabilizers are intersected as in~\eqref{eq:exploitable} accounting for their orientation. In Fig.~\ref{fig:cloth-flow}, $C_{2v}$ and $C_{4v}$ have perpendicular axes and only share the mirror $\Mx$, so that $\Gstar \!=\! C_s$. 

\section{Experiments}
\label{sec:experiments}

We study morphological symmetries along the manipulation pipeline, testing their preservation in a physics simulator and dynamics model, and leveraging them in a policy. The step map $f_{\Delta t}$ of the simulator or learned model advances the state under the external forces $\bm{\tau}_t$ of the robot(s) or hand so that $(\q_{t+1}, \dot\q_{t+1}) = f_{\Delta t}(\q_t, \dot\q_t, \bm \tau_t)$. 
For any $g \in \Gstar$ acting as in~\eqref{eq:act-disc}-\eqref{eq:actdisc-tancotan}, the \emph{equivariance error} at step $t$ is given by
\begin{equation}
\label{eq:equiv-error}
    \delta_t(g) = \big\Vert f_{\Delta t}\big(g \lgroupactionblank (\q_t, \dot\q_t, \bm{\tau}_t)\big) - g \lgroupactionblank f_{\Delta t}(\q_t, \dot\q_t, \bm{\tau}_t) \big\Vert.
\end{equation}
We report its root mean square over all the steps of a rollout, separately for positions and velocities.

\subsection{Symmetries in Simulators}
\label{subsec:experiments-simulators}
\begin{figure}[t]
    \centering
    \includegraphics[width=\linewidth]{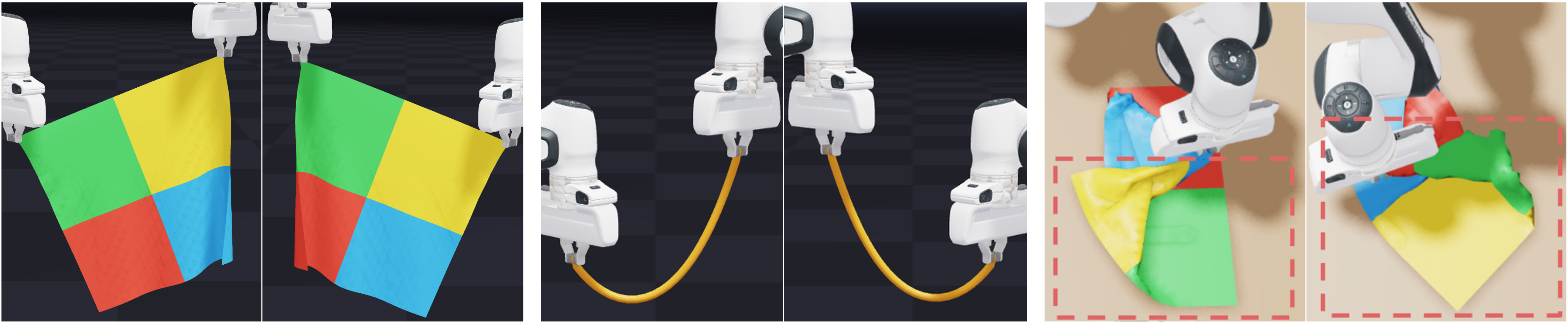}
    \caption{Mirrored cloth and rope lifting, and cloth folding in Newton.}
    \label{fig:sim_tasks_vis}
    \vspace{-0.4cm}
\end{figure}

First, we evaluate the extent to which a physics simulator preserves the exploitable symmetries of deformable objects. 
We consider three tasks where a cloth or a rope is manipulated by Franka Emika Panda FR3 robots in the Newton physics engine~\cite{newton}, see Fig.~\ref{fig:sim_tasks_vis}. In the lifting task, two robots hold a cloth by its top corners or a rope by its endpoints. One robot lifts the object vertically while the other stays still. In the folding task, a single robot grasps a cloth lying on a table by its left or right edge and folds it in half. A uniform, isotropic square cloth has $\Gmorph = D_{4h}$ and a uniform rope has $\Gmorph = D_{\infty h}$, which the environment and task reduce to $\Gstar = C_s = \{e, \Mx\}$, with $\Mx$ the mirror across the mid-plane between the two arms (lifting) or the vertical plane bisecting the cloth (folding).

Newton discretizes the object into a mesh of $N$ particles and advances its state $(\bm{q}_t, \dot{\bm{q}}_t)$ by iteratively solving an implicit Euler step subject to contact and kinematic grasp constraints. 
On a mirror-symmetric mesh, the discretized dynamics satisfy~\eqref{eq:eom-equivariance} for $g \!\in\! \Gstar$, so that any error $\delta_t(\Mx)$ in~\eqref{eq:equiv-error} arises from numerical approximations in the solver. 

Table~\ref{tab:sim_results} reports the equivariance error over pairs of mirrored runs for $5$ evenly-spaced lift heights between $0.05$ and $0.15\,\mathrm{m}$ for lifting, and $3$ grasp points at $\sfrac{1}{4}$, $\sfrac{1}{2}$, and $\sfrac{3}{4}$ of the edge length for folding. In the lifting task, the suspended object evolves smoothly and the mirror symmetry is nearly preserved, with errors below a millimeter. In contrast, in the folding task, non-smooth events such as contact activation and friction transitions amplify numerical asymmetries, raising the error to the order of centimeters. Therefore, the simulator preserves the exploitable symmetry only approximately, and arguably less so in contact-rich tasks.\looseness-1

\begin{table}[t]
    \centering
    \caption{Equivariance error of Newton's rollout trajectories.}
    \label{tab:sim_results}
    \footnotesize
    \begin{tabular}{@{}ll r@{${}\pm{}$}l r@{${}\pm{}$}l@{}}
        \toprule
        Task & Object & \multicolumn{2}{c}{$\bm{q}$ RMSE [mm]}
                      & \multicolumn{2}{c@{}}{$\dot{\bm{q}}$ RMSE [mm/s]} \\
        \midrule
        \multirow{2}{*}{Lifting} & Cloth & 0.928 & 0.123 & 2.302  & 0.364 \\
                                 & Rope  & 0.117 & 0.003 & 0.517 & 0.016 \\
        \midrule
        Folding & Cloth & 13.88 & 7.834 & 26.07 & 8.598 \\
        \bottomrule
    \end{tabular}
    \vspace{-0.3cm}
\end{table}

\subsection{Symmetries of Learned Dynamics Models}
\label{subsec:experiments-learned-dynamics}
Second, we evaluate the extent to which deformable object dynamics learned from videos preserve morphological symmetries. 
We consider three scenarios from PhysTwin~\cite{jiang2025phystwin}, namely puppet pushing, rope shrinking, and rectangular cloth lifting, with RGB-D observations of hand-object interactions. 
The puppet and the rope each have a single mirror, i.e., $C_s$, and the rectangular cloth two perpendicular ones, i.e., $C_{2v}$.

PhysTwin~\cite{jiang2025phystwin} represents a deformable object as a mass-spring graph of particles, and connects the point cloud of the human hand, which serves as the action input, to nearby particles via learnable springs.
As the permutation $\bm{P}_{\sigma_g}$ is difficult to determine for particles reconstructed from visual observations, we symmetrize the particle set by mirroring a subset of it, whose correspondences directly determine $\bm{P}_{\sigma_g}$.

\begin{figure}
    \centering
    \includegraphics[width=\linewidth]{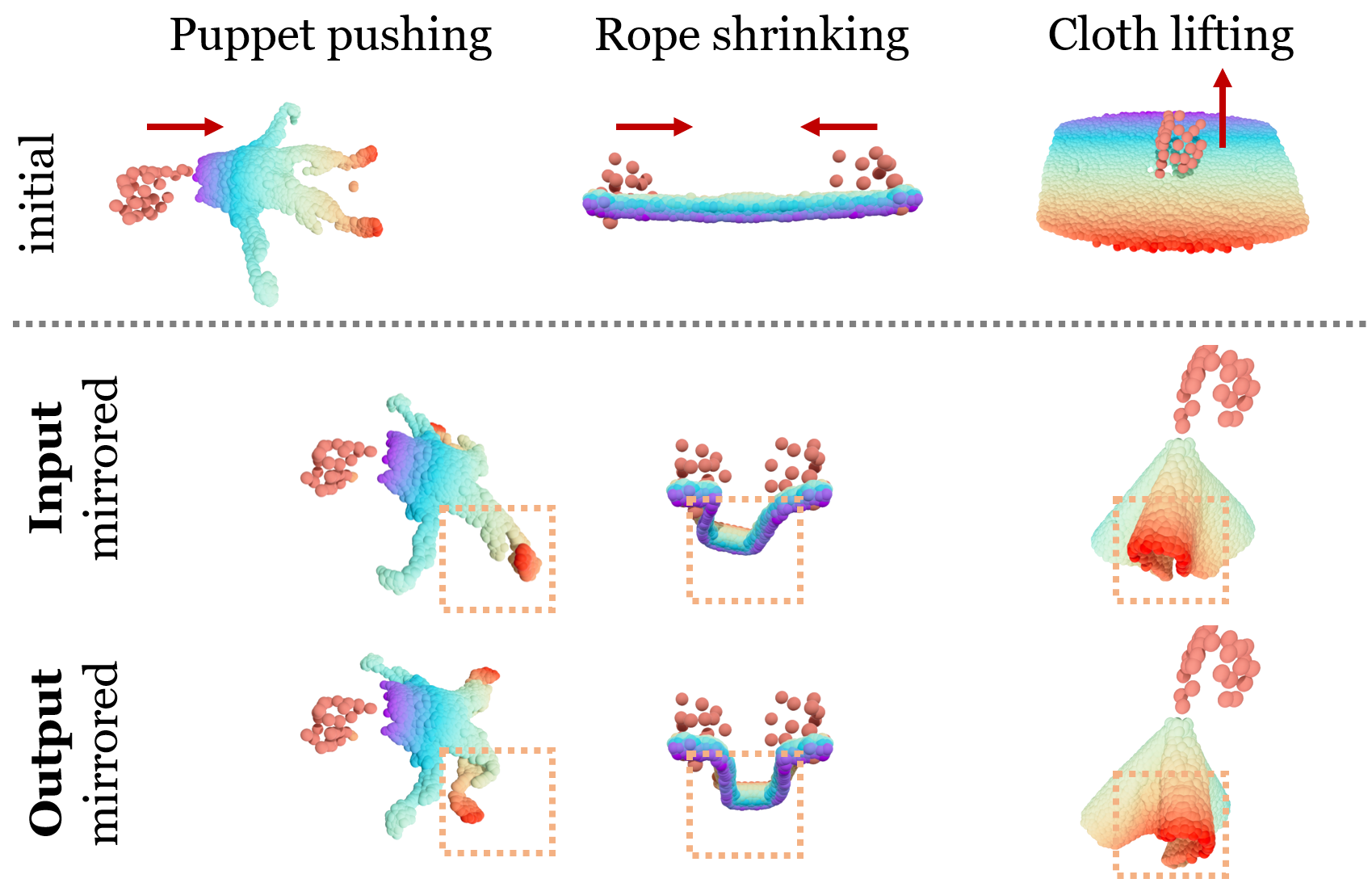}
    \caption{Mirrored PhysTwin rollouts. Particle colors indicate correspondences. Red points are the point cloud of the human hand (action input), and red arrows the direction of its motion.}
    \label{fig:phystwin_results}
    \vspace{-0.3cm}
\end{figure}

Fig.~\ref{fig:phystwin_results} compares the dynamics rollout $f_{\Delta t}\big(g \lgroupactionblank (\q_t, \dot\q_t, \bm{\tau}_t)\big)$ from mirrored input states and actions with the mirrored dynamics rollout 
$g \lgroupactionblank f_{\Delta t}(\q_t, \dot\q_t, \bm{\tau}_t)$.
The two arguably differ, particularly for the legs of the puppet and the folds of the cloth.
The equivariance errors~\eqref{eq:equiv-error}, evaluated over five trials with the same initial state and hand action and quantified in Table~\ref{tab:equivariance_error}, are generally larger than those in Table~\ref{tab:sim_results}, demonstrating that the learned dynamics model is not equivariant to morphological symmetries, i.e., learning the dynamics model from scratch yields asymmetric physical parameters.

\begin{table}[t]
\centering
\caption{Equivariance error of PhysTwin's rollout trajectories.}
\label{tab:equivariance_error}
\begin{tabular}{lcc}
\toprule
Task & $\boldsymbol{q}$ RMSE [mm] & $\dot{\boldsymbol{q}}$ RMSE [mm/s] \\
\midrule
Puppet pushing & $18.36_{\pm 5.10}$ & $52.76_{\pm 2.91}$ \\
Rope shrinking & $15.17_{\pm 0.00}$ & $59.85_{\pm 0.00}$ \\
Cloth lifting & $34.49_{\pm 2.82}$ & $50.63_{\pm 1.89}$ \\
\bottomrule
\end{tabular}
\vspace{-0.3cm}
\end{table}

\subsection{Symmetries of Policies}
\label{subsec:experiments-policies}
Finally, we show that a policy that exploits $\Gstar$ generalizes to symmetric scenes. We consider an RB-Y1 robot that folds a T-shirt lying on a table in half, grasping the side closest to its hands with both arms.
Assuming identical front and back sides, a homogeneous T-shirt has $\Gmorph \!=\! C_{2v}$, i.e., the mirror $\Mx$ in its sagittal plane, the flip $\Fl$ that exchanges its faces, and their product.
Gravity and the table exclude $\Fl$, while the task of folding one side onto the other, and the discretization are $\Mx$-invariant, hence $\Gstar = C_s = \{e, \Mx\}$.

We learn policies $\pi \!:\! \mathcal{S} \!\to\! \mathcal{A}$ mapping gripper poses $(\bm p, \bm R) \!\in\! \speceuclideangroup{3}$ and openings, and the centroid of the shirt's point cloud to gripper twists $(\bm v, \bm\omega)$ and openings.
The mirror $\Mx$ acts as $\Mx \lgroupactionon{\specialeucl{3}}{(\bm p, \bm R)} \!=\! (\bm{\Mx} \bm p, \bm{\Mx} \bm R \bm{\Mx})$ and 
$\Mx \lgroupactionon{\speceuclideanliealgebra{3}}{(\bm{v}, \bm{\omega})} \!=\! (\bm{\Mx} \bm{v}, -\bm{\Mx} \bm{\omega})$, with $\bm{\Mx} \!=\! \rep{\euclideanspace^3}(\Mx)$, analogously to~\eqref{eq:act-F}. 
As we exploit the morphological symmetry of the shirt and not of the robot, these actions apply to each hand independently, without permuting the arms.

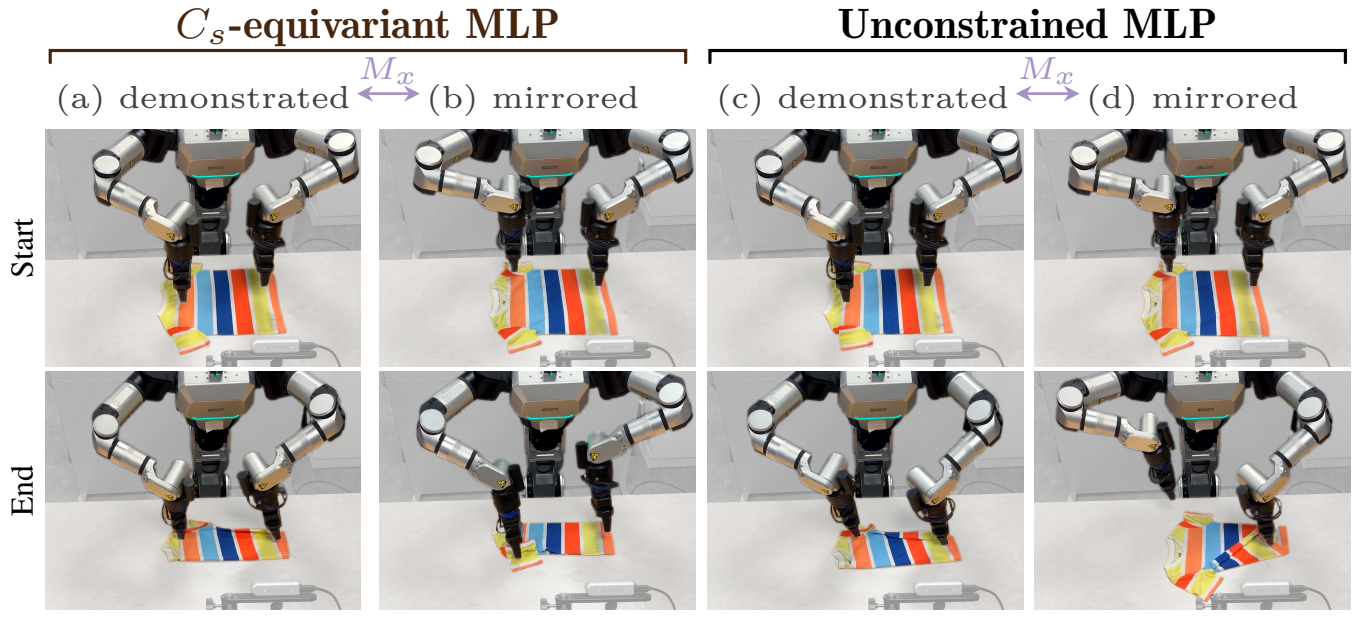
\begin{figure}[t]
    \centering
    \begin{tikzpicture}[x=\columnwidth, y=1cm,
        pol/.style={font=\fontfamily{lmr}\fontseries{b}\fontsize{8}{9}\selectfont, inner sep=1pt, anchor=base},
        brk/.style={line width=0.6pt, line cap=round},
        scn/.style={font=\fontfamily{lmr}\fontsize{6}{7}\selectfont, text=black!75, inner sep=0.5pt, anchor=base},
        mir/.style={{Stealth[length=3pt, width=3pt]}-{Stealth[length=3pt, width=3pt]}, line width=0.5pt, draw=nfPurple}]
        \foreach \l/\r/\c/\name in {0/0.4947/cflCstar/{\textcolor{cflCstar}{$C_s$}-equivariant MLP},
                                    0.5053/1/black/{Unconstrained MLP}
                                    } 
                                    {
            \draw[brk, draw=\c] (\l+0.004,0.30) -- (\l+0.004,0.38) -- (\r-0.004,0.38) -- (\r-0.004,0.30);
            \node[pol, text=\c] at ({(\l+\r)/2},0.46) {\name};
        }
        \foreach \a/\b/\la/\lb in {0.121/0.3737/a/b, 0.6263/0.879/c/d} {
            \node[scn] (o) at (\a,0) {(\la)~demonstrated};
            \node[scn] (m) at (\b,0) {(\lb)~mirrored};
            \path (o.base east) ++(0,0.6ex) coordinate (s) (m.base west) ++(0,0.6ex) coordinate (t);
            \draw[mir, shorten <=0.5pt, shorten >=0.5pt] (s) -- (t) node[midway, above=1.5pt, inner sep=0pt, font=\fontsize{6}{7}\selectfont, text=nfPurple] {$\Mx$};
        }
    \end{tikzpicture}\\[2pt]
\begin{subfigure}[t]{0.242\columnwidth}
    \begin{tikzpicture}[baseline=(start.south)]
        \node[inner sep=0pt, outer sep=0pt] (start) {
            \includegraphics[width=\linewidth]
            {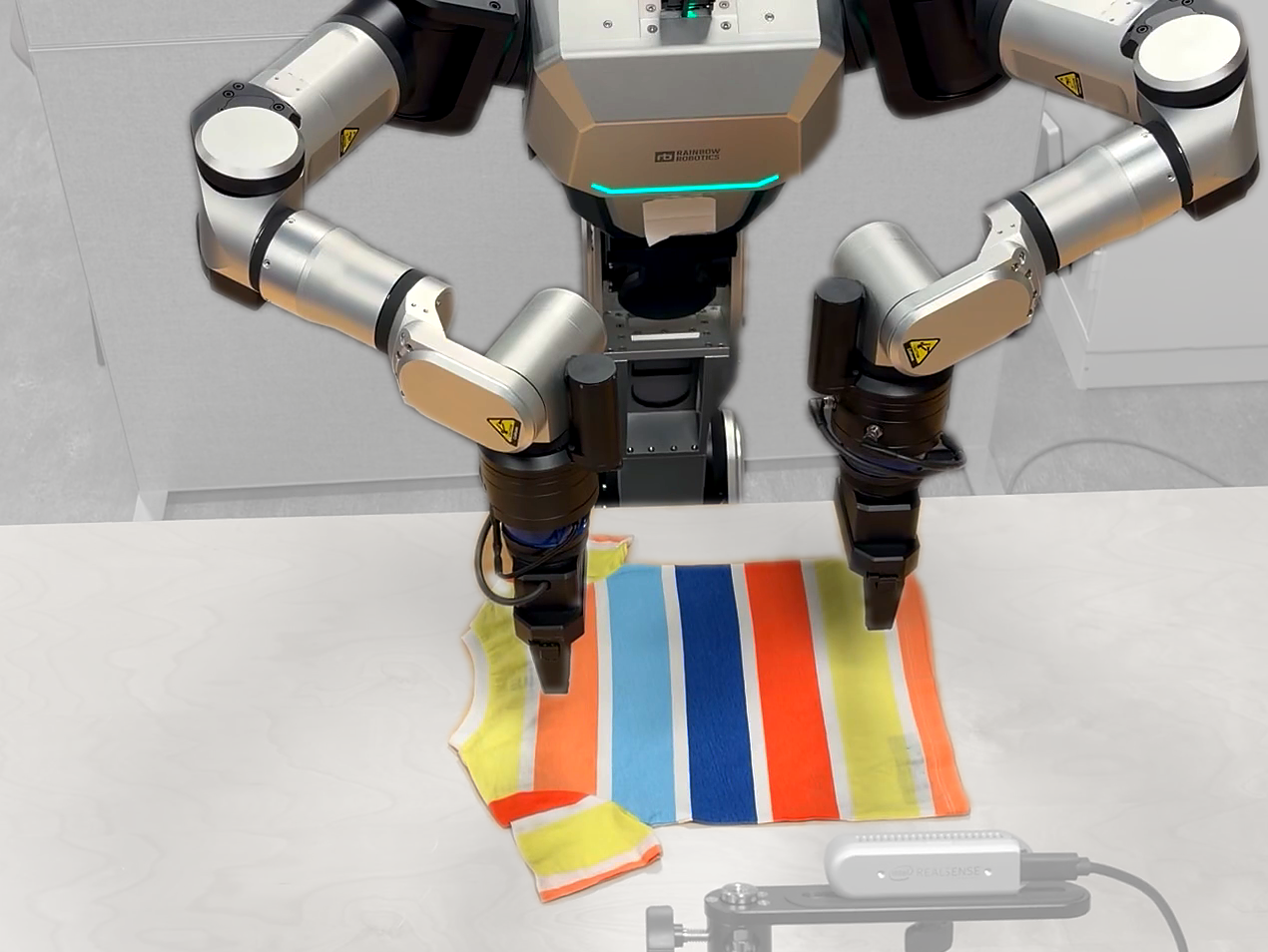}
        };
        \node[inner sep=0pt, outer sep=0pt, anchor=north]
            (end) at ([yshift=-1pt]start.south) {
            \includegraphics[width=\linewidth]
            {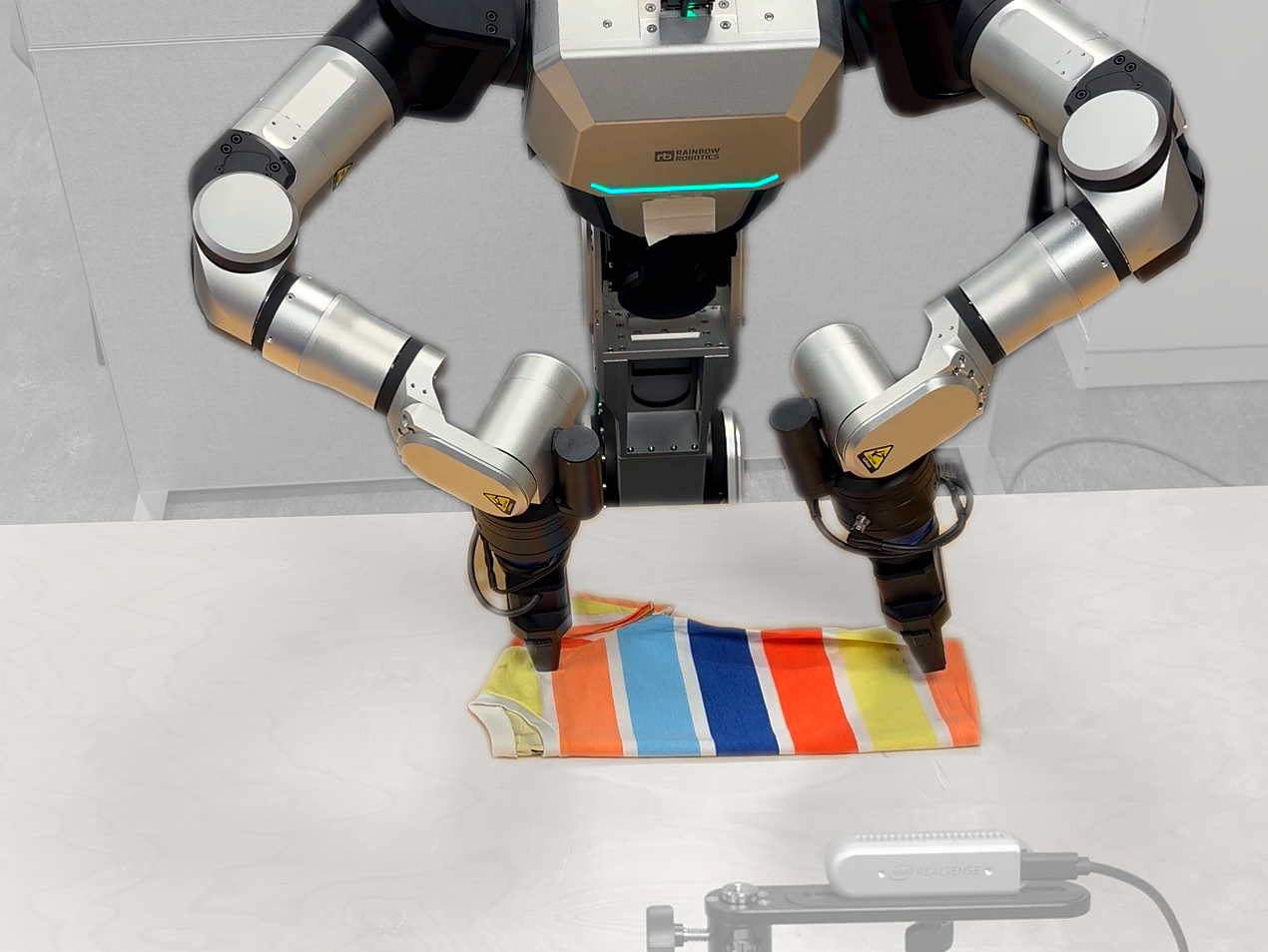}
        };
        \begin{scope}[overlay]
            \node[rotate=90, anchor=south, font=\fontsize{6}{7}\selectfont,
                  inner sep=2pt] at (start.west) {Start};
            \node[rotate=90, anchor=south, font=\fontsize{6}{7}\selectfont,
                  inner sep=2pt] at (end.west) {End};
        \end{scope}
    \end{tikzpicture}
    \phantomsubcaption\label{fig:policy_results_sym_orig}
\end{subfigure}    
    \begin{subfigure}[t]{0.242\columnwidth}
        \includegraphics[width=\linewidth]{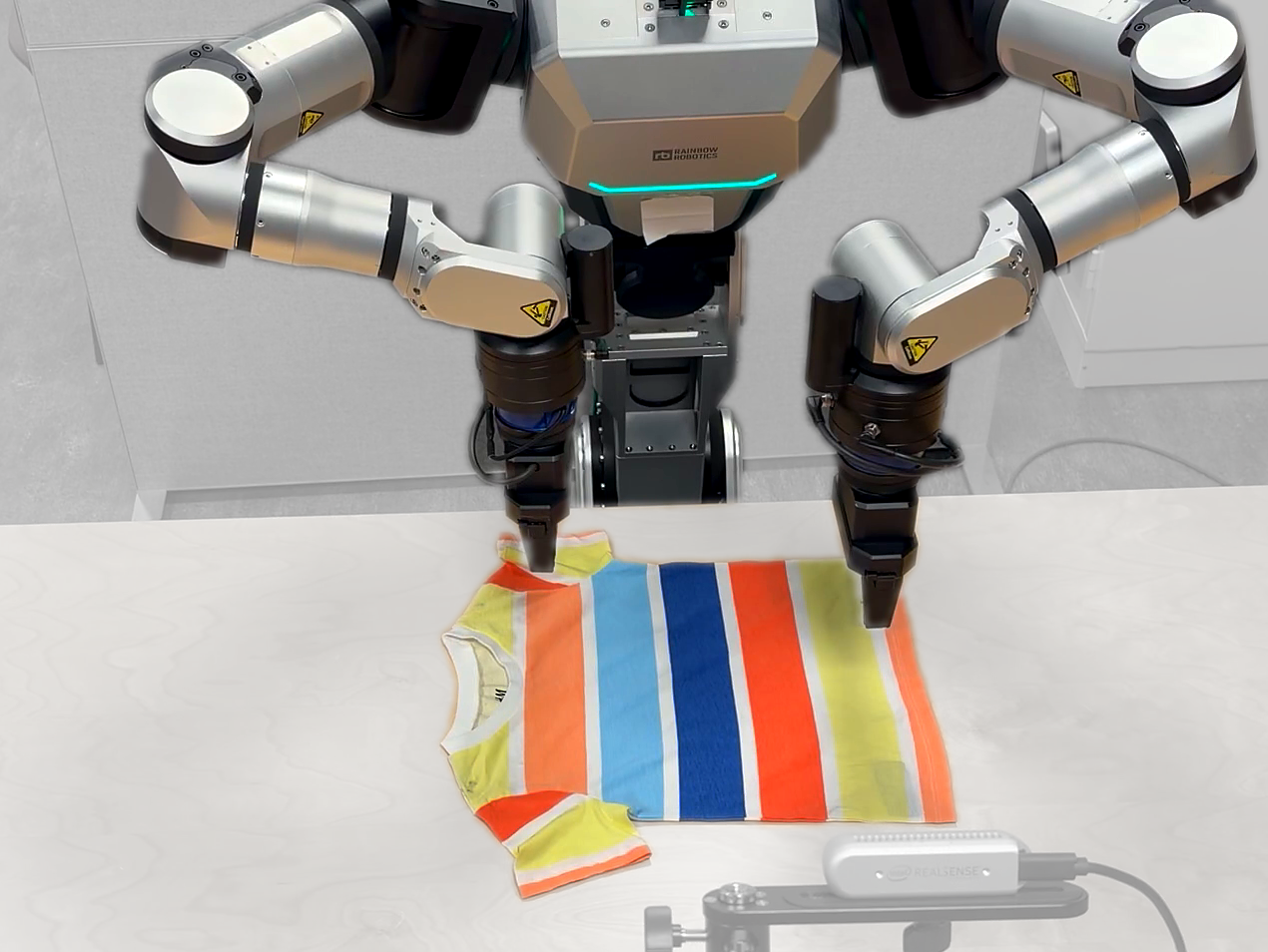}\\[1pt]
        \includegraphics[width=\linewidth]{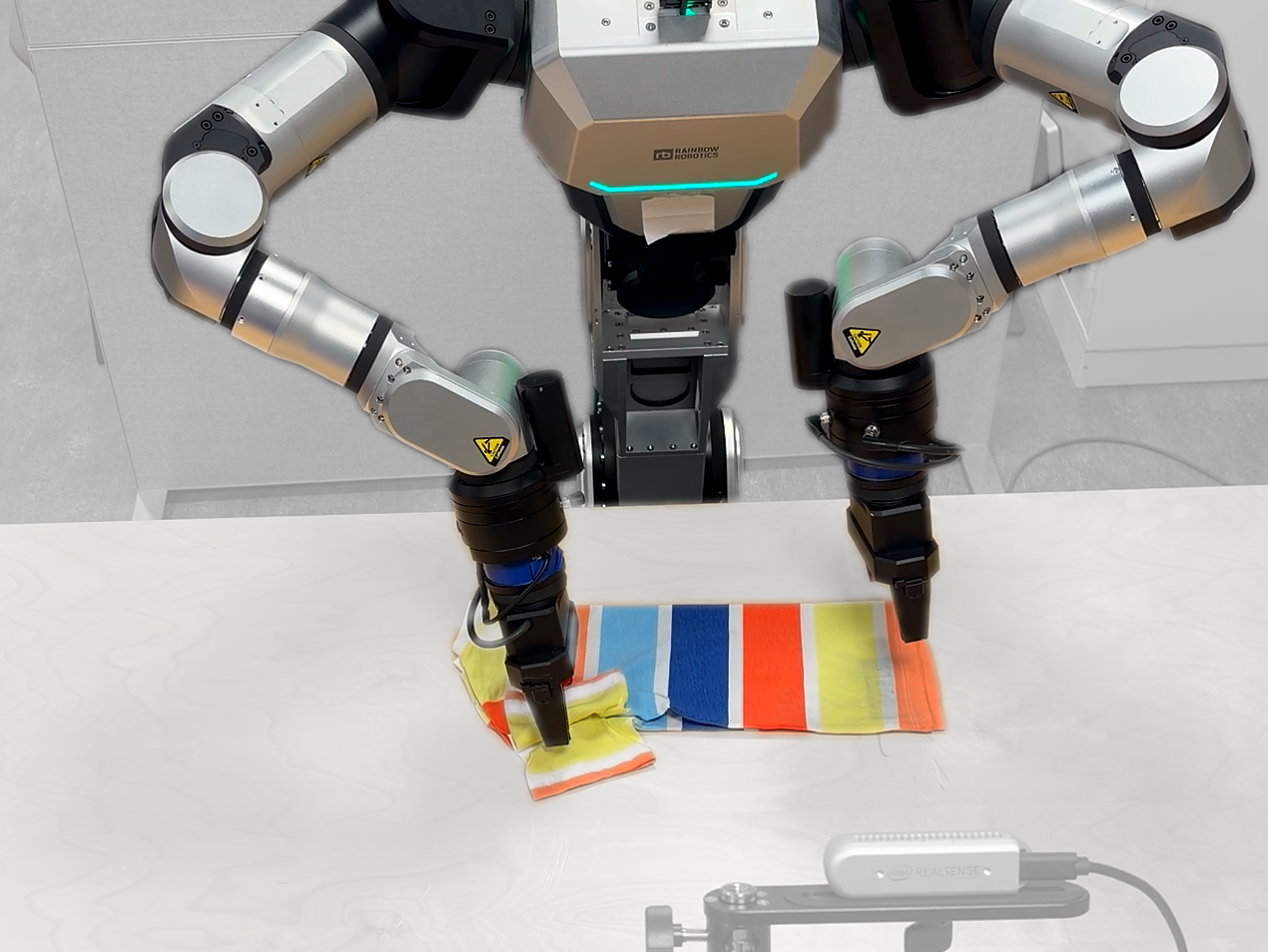}%
        \phantomsubcaption\label{fig:policy_results_sym_mirr}
    \end{subfigure}\hfill
    \begin{subfigure}[t]{0.242\columnwidth}
        \includegraphics[width=\linewidth]{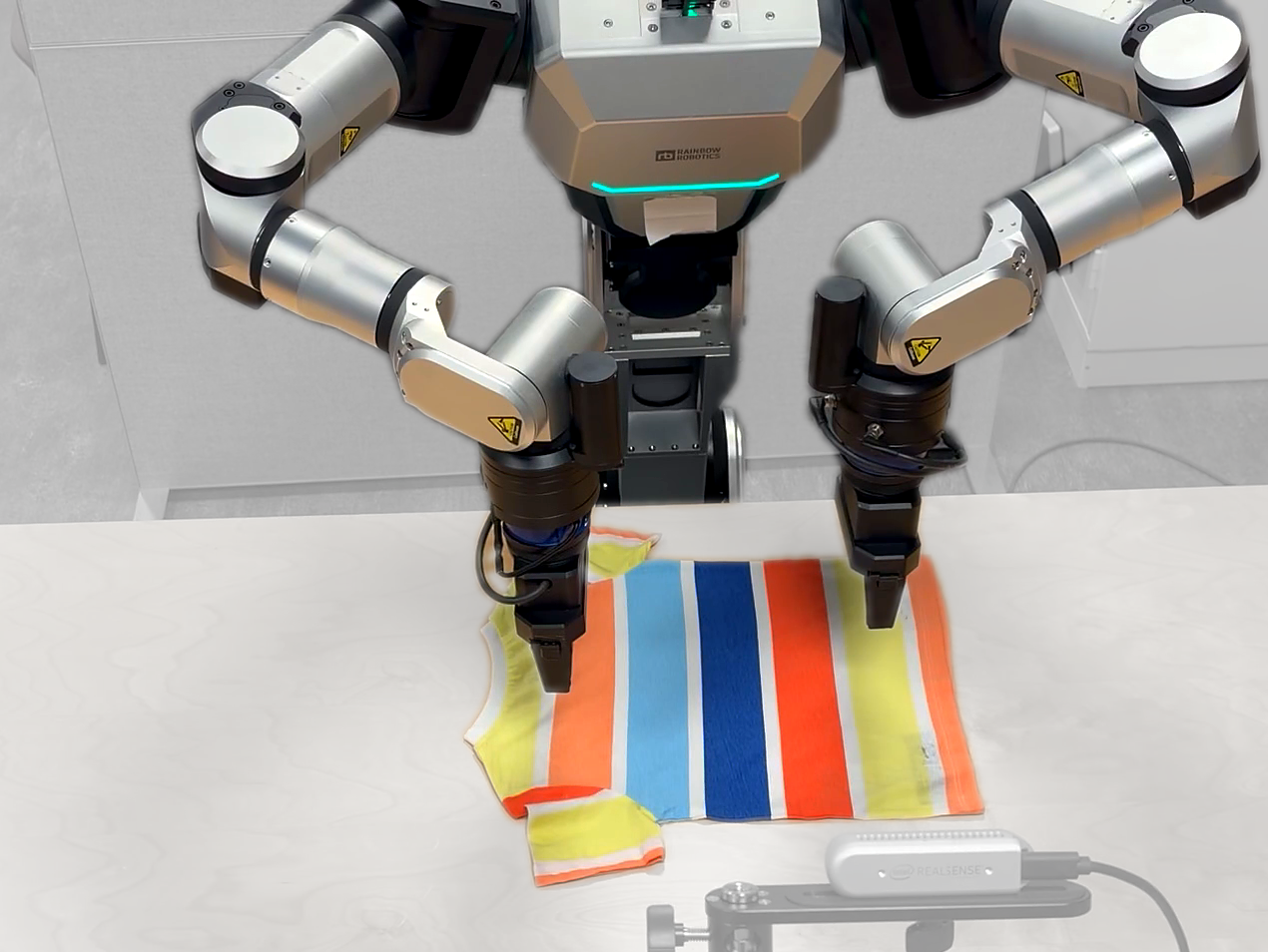}\\[1pt]
        \includegraphics[width=\linewidth]{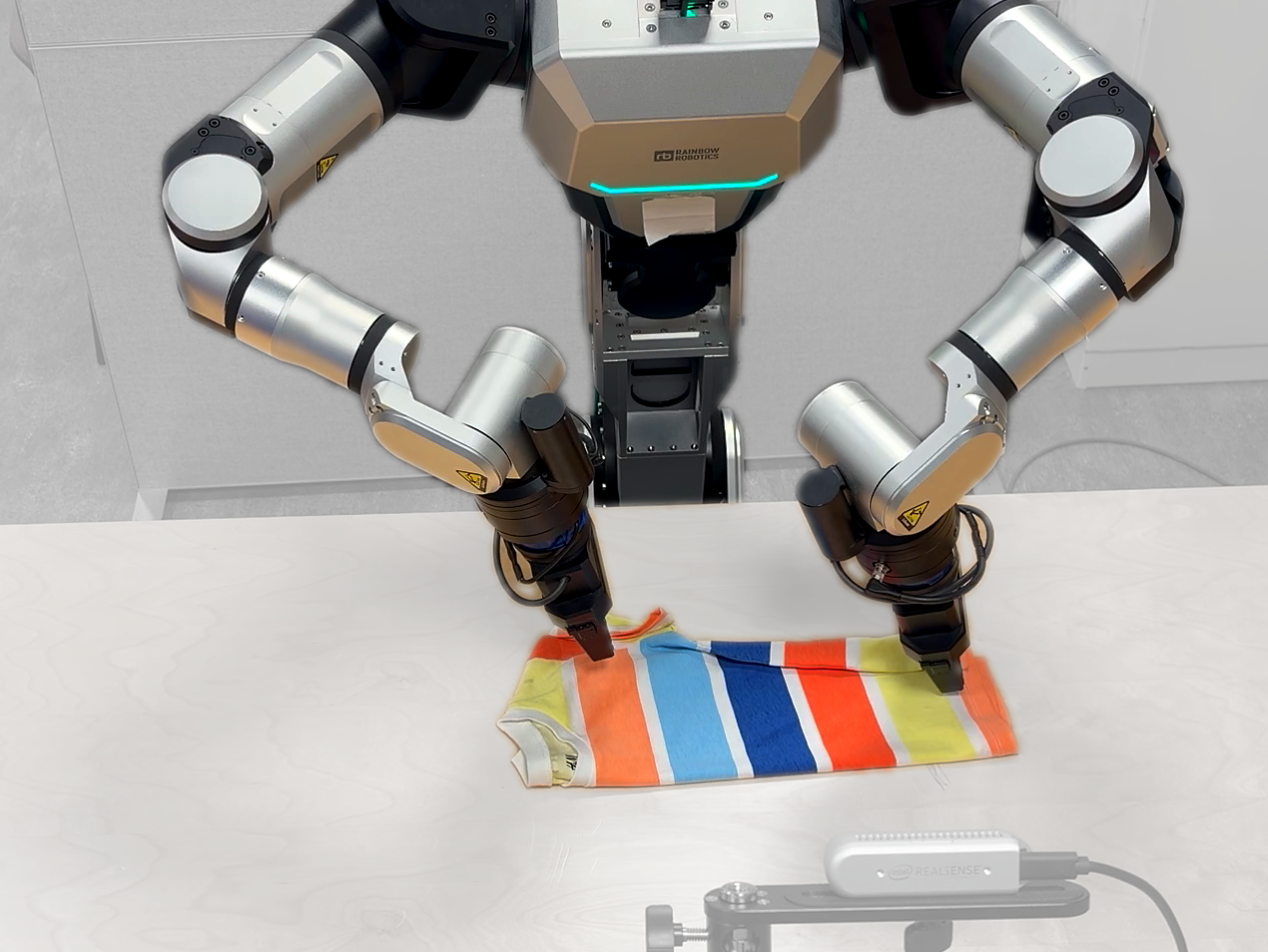}%
        \phantomsubcaption\label{fig:policy_results_plain_orig}
    \end{subfigure}\hfill
    \begin{subfigure}[t]{0.242\columnwidth}
        \includegraphics[width=\linewidth]{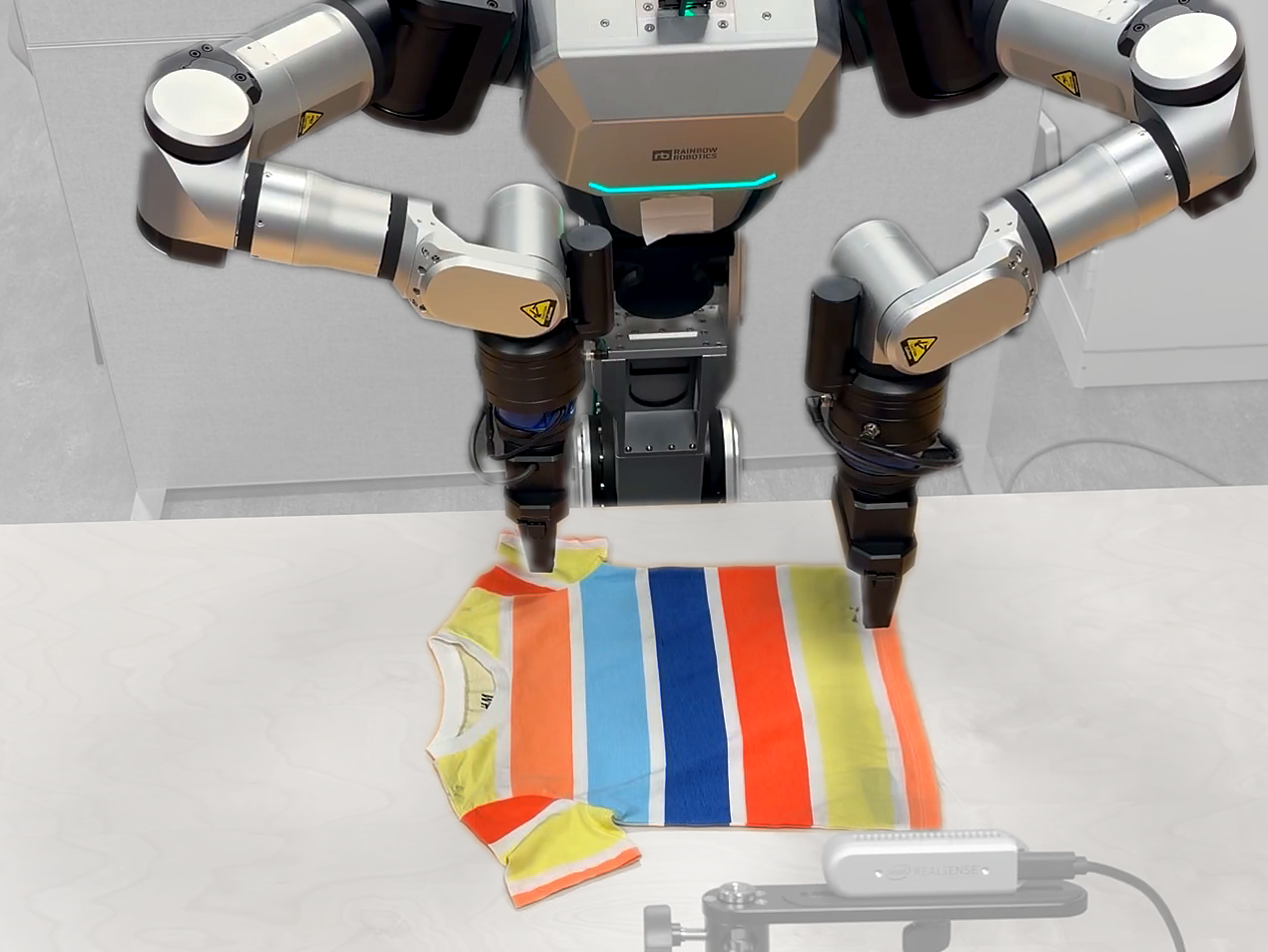}\\[1pt]
        \includegraphics[width=\linewidth]{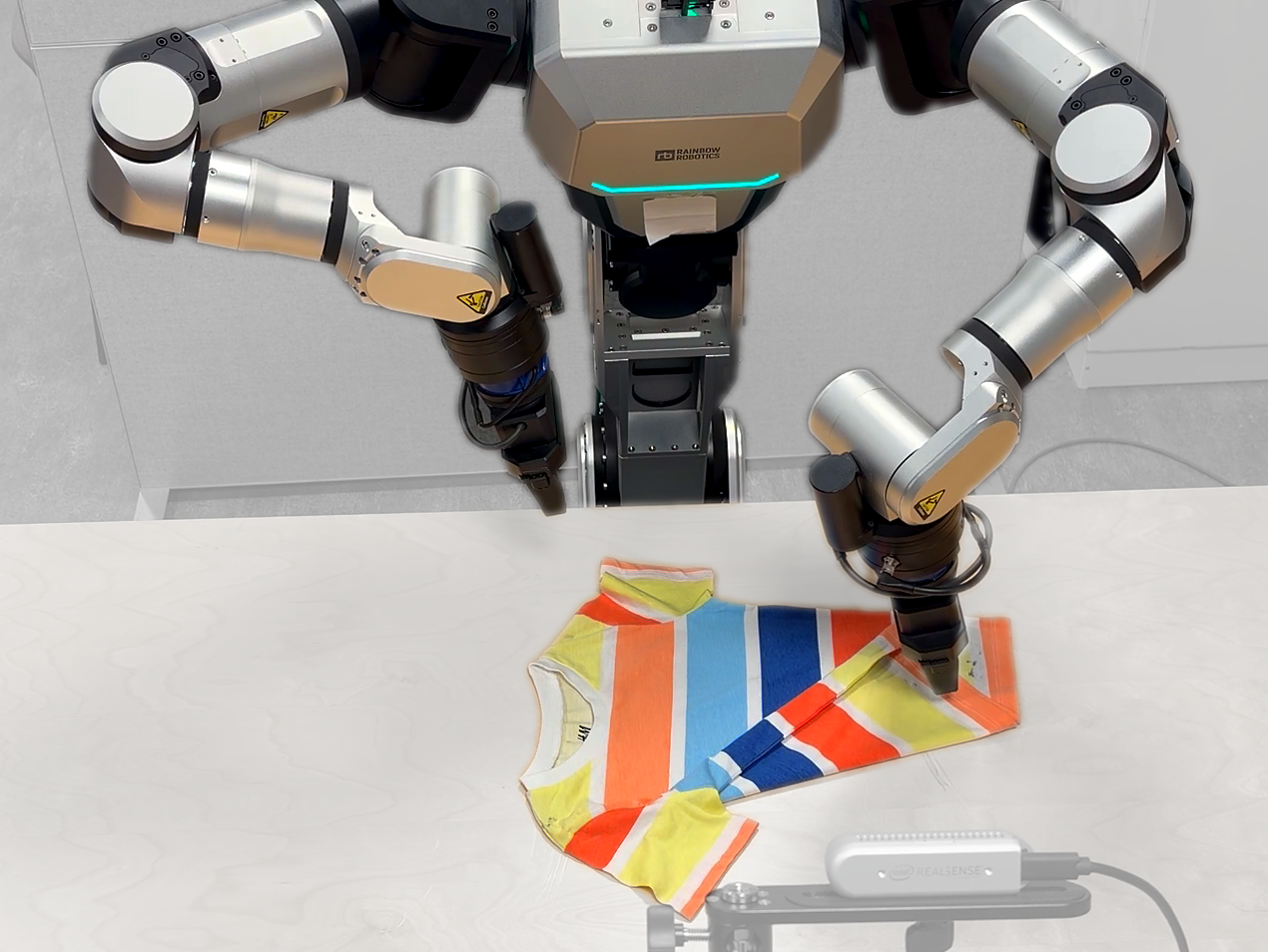}%
        \phantomsubcaption\label{fig:policy_results_plain_mirr}
    \end{subfigure}
    \caption{Rollouts of $C_s$-equivariant (a,\,b) and unconstrained (c,\,d) policies in the demonstrated (a,\,c) and $\Mx$-mirrored (b,\,d) scenes.}
    \label{fig:policy_results}
    \vspace{-0.3cm}
\end{figure}

We compare an unconstrained MLP policy $\pi$ with its frame average $\bar\pi(\bm s) = \frac{1}{|\Gstar|} \sum_{g \in \Gstar} g \lgroupactionon{\mathcal A}{\pi( g^{-1} \lgroupactionon{\mathcal S}{\bm s})}$~\cite{puny2022frame}, $\Gstar$-equivariant by construction.  
Both are trained on $12$ demonstrations and evaluated over $10$ trials in the demonstrated and mirrored scenes. 
Fig.~\ref{fig:policy_results} shows rollouts of both policies in the two scenes, and Table~\ref{tab:cloth_results} reports their success rates and the RMSE of the hand trajectories with respect to the (mirrored) mean demonstration. 
Both policies succeed in all trials in the demonstrated scene.
In the mirrored scene, $\bar\pi$ succeeds in all trials, generalizing without additional data, whereas $\pi$ consistently fails.
As $\bar\pi$ is exactly equivariant, its higher RMSE in the mirrored scene stems from an experimental setup that is only approximately symmetric.

\begin{table}[t]
    \centering
    \caption{Success rates and trajectory RMSE to the mean demo.}
    \label{tab:cloth_results}
    \footnotesize
    \begin{tabular}{llcc}
        \toprule
        Policy & Setup & Success & RMSE [mm] \\
        \midrule
        $C_s$-equivariant $\bar{\pi}(\bm{s})$ & Demonstrated & $10/10$ & $2.6_{\pm 0.6}$ \\
        $C_s$-equivariant $\bar{\pi}(\bm{s})$ & Mirrored & $10/10$ & $6.3_{\pm 0.9}$ \\
        \midrule
        Unconstrained $\pi(\bm{s})$ & Demonstrated & $10/10$ & $2.0_{\pm 0.3}$ \\
        Unconstrained $\pi(\bm{s})$ & Mirrored & $0/10$ & -- \\
        \bottomrule
    \end{tabular}
     \vspace{-0.2cm}
\end{table}

\section{Discussion}
\label{sec:discussion}

This paper formalized the morphological symmetries of deformable objects, and introduced an intersection principle and a lattice-based taxonomy to systematically identify the exploitable symmetry group of a manipulation problem.
Our experiments showed that simulators and learned dynamics models do not generally preserve these symmetries, while policies leveraging them generalize to symmetric scenes.

The intersection principle has two practical consequences.
First, as each constituent can only remove symmetries, $\Gstar$ is typically much smaller than the stabilizer of the object.
However, even a group of two elements suffices for a policy to generalize to a mirrored scene without additional data (Sec.~\ref{subsec:experiments-policies}).
Second, a model that exploits $\Gstar$ is tied to the environment and task for which $\Gstar$ was identified.
This motivates the development of independent dynamics models that exploit $\Gmorph \cap \Gdisc$ along with strategies to combine them with the symmetries of each environment and task.

In practice, symmetries rarely hold exactly, e.g., due to seams, an off-grain weave, a tilted table, or sensor noise.
Following~\cite{petrache2023approximation-generalization-ad6}, we measure the violation of a symmetry $g$ by a constituent $x$ with its relative defect ${\delta_x(g) \!=\! \Vert g \lgroupactionon{\mathcal X}{x} - x \Vert / \Vert x \Vert}$, and collect the elements with a defect of at most $\epsilon$ in its $\epsilon$-stabilizer $\group_x^{\epsilon}$.
As defects add up under composition, $\group_x^{\epsilon}$ is not a group in general, and we define an $\epsilon$-approximate exploitable group $\Gstar_\epsilon$ as a largest subgroup of $\orthogonalgroup{3}$ contained in all $\group_x^{\epsilon}$. Such a subgroup need not be unique, but typically, one of them coincides with the exact group $\Gstar$ of an idealized problem that neglects these imperfections, to which the intersection principle and the taxonomy then apply directly.
Exploiting it reduces the generalization error of a model, but introduces an error of order $\epsilon$, which propagates to the dynamics, where it accumulates over a rollout, and to the value function of the task~\cite{petrache2023approximation-generalization-ad6, park2024approximate-319}.
A symmetry that holds up to a small systematic defect can thus still be exploited to improve the generalization of dynamics models and policies.

\bibliographystyle{IEEEtran}
\bibliography{References.bib}

\clearpage

\useRomanappendicesfalse
\appendices
\setcounter{equation}{0}\renewcommand{\theequation}{S\arabic{equation}}\renewcommand{\theHequation}{S\arabic{equation}}%
\setcounter{proposition}{0}\renewcommand{\theproposition}{S\arabic{proposition}}\renewcommand{\theHproposition}{S\arabic{proposition}}%
\setcounter{table}{0}\renewcommand{\thetable}{S\arabic{table}}\renewcommand{\theHtable}{S\arabic{table}}%
\section{Proofs of Section~\ref{sec:morphological_symmetry}}
\label{app:proofs}

Throughout, we use that, for $g \in \Gshape$, the substitution $\bm X = g \lgroupactionblank \bm Y$ maps $\body$ onto itself with $\mathrm d\bm X = \mathrm d\bm Y$, and that $\rep{\euclideanspace^3}(g)$ is orthogonal, i.e., $\rep{\euclideanspace^3}^{-1}(g) = \rep{\euclideanspace^3}^{\trsp}(g)$.

\begin{proof}[Proof of Proposition~\ref{prop:perterm}]
Let $\varphi' = g \lgroupactionon{\configmanifold}{\varphi}$, with velocity $\dot\varphi'(\bm X) = g \lgroupactionblank \dot\varphi(g^{-1} \lgroupactionblank \bm X)$ and deformation gradient $\bm F'(\bm X) = \rep{\euclideanspace^3}(g) \, \bm F(g^{-1} \lgroupactionblank \bm X) \, \rep{\euclideanspace^3}^{\trsp}(g)$ by~\eqref{eq:act-tang} and~\eqref{eq:act-F}. Substituting $\bm X = g \lgroupactionblank \bm Y$ in~\eqref{eq:energies} and using $|g \lgroupactionblank \dot\varphi| = |\dot\varphi|$ yields
\begin{align}
    T(\dot\varphi') &= \frac12 \int_{\body} \rho(g \lgroupactionblank \bm Y) \, |\dot\varphi(\bm Y)|^2 \, \mathrm d\bm Y,
    \label{eq:proof-kin} \\
    V_{\mathrm{el}}(\varphi') &= \int_{\body} W\big(g \lgroupactionblank \bm Y, \, \rep{\euclideanspace^3}(g) \bm F(\bm Y) \rep{\euclideanspace^3}^{\trsp}(g)\big) \, \mathrm d\bm Y,
    \label{eq:proof-el} \\
    V_{\mathrm{gr}}(\varphi') &= - \int_{\body} \rho(g \lgroupactionblank \bm Y) \, (g^{-1} \lgroupactionblank \gvec) \cdot \varphi(\bm Y) \, \mathrm d\bm Y .
    \label{eq:proof-gr}
\end{align}
Compared to the energies of $\varphi$, $\rho$ is replaced by $\rgroupactionon{}{\rho} \, g$ in~\eqref{eq:proof-kin}, $W$ by $\rgroupactionon{}{W} \, g$ in~\eqref{eq:proof-el}, and $\rho \, \gvec$ by $(\rgroupactionon{}{\rho} \, g) \, (g^{-1} \lgroupactionblank \gvec)$ in~\eqref{eq:proof-gr}. They are thus unchanged for $g \in \Gmass$, $g \in \Gmat$, and $g \in \Gmass \cap \Ggrav$, respectively, and $\mathcal L$ on the intersection~\eqref{eq:G_lagrangian}.

The conditions are also necessary for $T$ and $V_{\mathrm{gr}}$: as $\dot\varphi$ and $\varphi$ are arbitrary, invariance requires $\rgroupactionon{}{\rho} \, g = \rho$ and $(\rgroupactionon{}{\rho} \, g) \, (g^{-1} \lgroupactionblank \gvec) = \rho \, \gvec$ pointwise, and taking norms with $\rho > 0$ and $\gvec \neq \bm 0$ splits the latter into $g \in \Gmass$ and $g \lgroupactionblank \gvec = \gvec$. For $V_{\mathrm{el}}$, $\rgroupactionon{}{W} \, g = W$ is necessary up to null Lagrangians.
\end{proof}

\begin{proof}[Proof of Proposition~\ref{prop:equivariance}]
A motion $\varphi(t)$ solves~\eqref{eq:elastodynamics} with the boundary conditions $\bm P \bm n = \bm t$ on $\partial\body_{\mathrm N}$ and $\varphi = \varphi_{\mathrm D}$ on $\partial\body_{\mathrm D}$ iff it satisfies the latter and the Lagrange--d'Alembert principle
\begin{equation}
\label{eq:lda}
    \delta \! \int \! \mathcal L(\varphi, \dot\varphi) \, \mathrm dt + \! \int \! \Big( \langle b, \delta\varphi \rangle + \! \int_{\partial\body_{\mathrm N}} \! \bm t \cdot \delta\varphi \, \mathrm dA \Big) \mathrm dt = 0
\end{equation}
for all variations $\delta\varphi$ that vanish on $\partial\body_{\mathrm D}$ and at the end times~\cite{marsden1983mathematical-553}. Let $\varphi' = g \lgroupactionon{\configmanifold}{\varphi}$, $b' = g \lgroupactionblank b$, and $\bm t' = g \lgroupactionblank \bm t$ as in~\eqref{eq:act-cotang}. As the action~\eqref{eq:act-cont} is linear and invertible, the variations of $\varphi'$ are exactly the fields $\delta\varphi' = g \lgroupactionblank \delta\varphi$, and $\delta\varphi'$ vanishes on $g \lgroupactionblank \partial\body_{\mathrm D}$ iff $\delta\varphi$ vanishes on $\partial\body_{\mathrm D}$. We compare the terms of~\eqref{eq:lda} for $(\varphi', \delta\varphi')$ and $(\varphi, \delta\varphi)$:
The first terms coincide for $g \in \group_{\mathcal L}$.
With $\bm X = g \lgroupactionblank \bm Y$ and the orthogonality of $\rep{\euclideanspace^3}(g)$, $\langle b', \delta\varphi' \rangle = \int_{\body} (g \lgroupactionblank b(\bm Y)) \cdot (g \lgroupactionblank \delta\varphi(\bm Y)) \, \mathrm d\bm Y = \langle b, \delta\varphi \rangle$.
The traction term over $g \lgroupactionblank \partial\body_{\mathrm N}$ coincides analogously, as $g$ maps $\partial\body$ onto itself and preserves the area element $\mathrm dA$.
Hence~\eqref{eq:lda} holds for $\varphi'$ with $b'$ and $\bm t'$ on $g \lgroupactionblank \partial\body_{\mathrm N}$, and, by~\eqref{eq:act-cont}, $\varphi'$ takes the boundary values $g \lgroupactionon{\configmanifold}{\varphi_{\mathrm D}}$ on $g \lgroupactionblank \partial\body_{\mathrm D}$. This is the continuum counterpart of~\cite[Eq.~(9)]{apraez2025morphological}.
\end{proof}
\vspace{-2pt}
\begin{proposition}[Spatial--material symmetries]
\label{prop:pairs}
Let $u \in \euclideangroup{3}$, $u \lgroupactionblank \bm x = \bm Q_u \bm x + \bm d_u$, and $h \in \Gshape$ with $\det \bm Q_u = \det \rep{\euclideanspace^3}(h)$. Then
\begin{equation}
\label{eq:act-pair-app}
    (u, h) \lgroupactionon{\configmanifold}{\varphi} = u \circ \varphi \circ h^{-1}
\end{equation}
is an admissible configuration and defines a left action of these pairs on $\configmanifold$. If $h \in \Gmorph$ and $\bm Q_u \gvec = \gvec$, it changes $\mathcal L$ only by a constant and thus maps solutions onto solutions.
\end{proposition}
\begin{proof}
Let $\varphi' = u \circ \varphi \circ h^{-1}$ and $\bm R = \bm Q_u \rep{\euclideanspace^3}^{\trsp}(h)$ $\in \specorthogonalgroup{3}$, as the determinants are equal.
\emph{Admissibility and action:} The deformation gradient of $\varphi'$ is $\bm F'(\bm X) = \bm Q_u \, \bm F(h^{-1} \lgroupactionblank \bm X) \, \rep{\euclideanspace^3}^{\trsp}(h)$, with determinant $\det \bm R \, \det \bm F > 0$. Moreover, $(u_1, h_1) \lgroupactiononblank{\configmanifold} \big((u_2, h_2) \lgroupactiononblank{\configmanifold} \varphi\big) = (u_1 \circ u_2) \circ \varphi \circ (h_1 \circ h_2)^{-1}$, and pairs with equal determinants are closed under this product.

\emph{Lagrangian:} The velocity of $\varphi'$ is $\bm Q_u \, \dot\varphi \circ h^{-1}$ and $\bm F' = \bm R \, \rep{\euclideanspace^3}(h) \, (\bm F \circ h^{-1}) \, \rep{\euclideanspace^3}^{\trsp}(h)$. By frame indifference, $\bm R$ does not change the elastic energy density. Substituting $\bm X = h \lgroupactionblank \bm Y$ as in the proof of Proposition~\ref{prop:perterm}, $T$ and $V_{\mathrm{el}}$ thus take the forms~\eqref{eq:proof-kin} and~\eqref{eq:proof-el} with $h$ in place of $g$, and are unchanged for $h \in \Gmass$ and $h \in \Gmat$. For $h \in \Gmass$ and $\bm Q_u \gvec = \gvec$, the gravitational energy becomes
\begin{align}
    V_{\mathrm{gr}}(\varphi') &= -\int_{\body} (\rgroupactionon{}{\rho} \, h)(\bm Y) \; \gvec \cdot \big(\bm Q_u \varphi(\bm Y) + \bm d_u\big) \, \mathrm d\bm Y \nonumber \\
    &= V_{\mathrm{gr}}(\varphi) - m \, \gvec \cdot \bm d_u ,
\end{align}
with $m$ the mass of the body. As a constant offset of $\mathcal L$ does not affect the equations of motion, solutions are mapped onto solutions, with forces and boundary conditions transformed as in Proposition~\ref{prop:equivariance}.
\end{proof}

\begin{proof}[Proof of Proposition~\ref{prop:disc-action}]
\emph{Action:} Evaluating~\eqref{eq:act-cont} at node $\sigma_g(i)$ and using~\eqref{eq:Gdisc} gives
\begin{equation}
    (g \lgroupactionon{\configmanifold}{\varphi})(\bm X_{\sigma_g(i)}) = g \lgroupactionblank \varphi(g^{-1} \lgroupactionblank \bm X_{\sigma_g(i)}) = \rep{\euclideanspace^3}(g) \, \varphi(\bm X_i),
\end{equation}
i.e., node $\sigma_g(i)$ receives the transformed position of node $i$, which is~\eqref{eq:act-disc} with $(\bm P_{\sigma_g})_{ji} = \delta_{j \sigma_g(i)}$.

\emph{Representation:} As $(g_1 \circ g_2) \lgroupactionblank \bm X_i \!=\! g_1 \lgroupactionblank \bm X_{\sigma_{g_2}(i)} \!=\! \bm X_{\sigma_{g_1}(\sigma_{g_2}(i))}$, we have $\sigma_{g_1 \circ g_2} \!=\! \sigma_{g_1} \circ \sigma_{g_2}$ and thus $\bm P_{\sigma_{g_1}} \bm P_{\sigma_{g_2}} \!=\! \bm P_{\sigma_{g_1 \circ g_2}}$. Together with $\rep{\euclideanspace^3}(g_1) \rep{\euclideanspace^3}(g_2) \!=\! \rep{\euclideanspace^3}(g_1 \circ g_2)$, the mixed-product property of the Kronecker product gives $\rep{\redmanifold}(g_1) \rep{\redmanifold}(g_2) \!=\! \rep{\redmanifold}(g_1 \circ g_2)$.

\emph{Orthogonality and lifts:} $\rep{\redmanifold}(g)$ is orthogonal as $\bm P_{\sigma_g}$ and $\rep{\euclideanspace^3}(g)$ are. As~\eqref{eq:act-disc} is linear, velocities transform by $\rep{\redmanifold}(g)$, and generalized forces by $\rep{\redmanifold}^{-\trsp}(g) = \rep{\redmanifold}(g)$, which preserves the power $\bm\tau^\trsp \dot\q$.
\end{proof}

\begin{proof}[Proof of Proposition~\ref{prop:equivariance-disc}]
We abbreviate $\q' = g \lgroupactionon{\redmanifold}{\q}$ and $\dot\q' = g \lgroupactionon{\redtangentspace{\q}}{\dot\q}$.
\emph{Invariance:} Setting $\dot\q = \bm 0$ shows that $\mathcal L = T - V$ is invariant iff $V$ and $T$ are. By~\eqref{eq:act-disc}, the transformed kinetic energy is
\begin{equation}
    T(\q', \dot\q') = \tfrac12 \, \dot\q^\trsp \, \rep{\redmanifold}^\trsp(g) \, \bm M(\q') \, \rep{\redmanifold}(g) \, \dot\q .
\end{equation}
As the matrices are symmetric, it equals $T(\q, \dot\q)$ for all $\dot\q$ iff $\rep{\redmanifold}^\trsp(g) \bm M(\q') \rep{\redmanifold}(g) = \bm M(\q)$, which, with $\rep{\redmanifold}^{-1}(g) = \rep{\redmanifold}^\trsp(g)$, is~\eqref{eq:disc-massinvariance}, as in~\cite{apraez2025morphological}.

\emph{Equivariance:} Differentiating $\mathcal L(\q', \dot\q') = \mathcal L(\q, \dot\q)$ with respect to $\q$ gives
\begin{equation}
    \frac{\partial \mathcal L}{\partial \q}(\q, \dot\q) = \rep{\redmanifold}^\trsp(g) \, \frac{\partial \mathcal L}{\partial \q}(\q', \dot\q'),
\end{equation}
and analogously for $\dot\q$. If $\q(t)$ solves the Euler--Lagrange equations $\frac{\mathrm d}{\mathrm dt} \frac{\partial \mathcal L}{\partial \dot\q} - \frac{\partial \mathcal L}{\partial \q} = \bm\tau$, the transformed trajectory $\q'(t)$ thus solves them with the force $\rep{\redmanifold}^{-\trsp}(g) \bm\tau = g \lgroupactionon{\redcotangentspace{\q}}{\bm\tau}$. Its acceleration is $\rep{\redmanifold}(g) \, \ddot\q$, which is~\eqref{eq:eom-equivariance}.
\end{proof}

\end{document}